%% file: camera_ready.tex
\documentclass[sigconf]{acmart}

\copyrightyear{2026}
\acmYear{2026}
\setcopyright{cc}
\setcctype{by}
\acmConference[WWW '26] {Proceedings of the ACM Web Conference 2026}{April 13--17, 2026}{Dubai, United Arab Emirates.}
\acmBooktitle{Proceedings of the ACM Web Conference 2026 (WWW '26), April 13--17, 2026, Dubai, United Arab Emirates}
\acmISBN{979-8-4007-2307-0/2026/04}
\acmDOI{10.1145/XXXXXX.XXXXXX}

\usepackage{enumitem}
\usepackage{url}

\usepackage{balance}
\usepackage{booktabs} 
\usepackage{algpseudocode}

\usepackage{tcolorbox}

\usepackage{subcaption}
\usepackage{graphicx}

\usepackage{algorithm}
\usepackage{amsfonts} 
\usepackage{amsmath} 
\usepackage{amsthm}
\newtheorem{defi}{Definition}

\newtheorem{rem}{Remark}

\newtheorem{example}{Example}

\usepackage{amsmath}
\usepackage{bm}
\usepackage{xspace}
\usepackage{xcolor}

\usepackage{multirow}
\usepackage{caption}

\tcbuselibrary{listings, skins, breakable}
\definecolor{lightgray}{gray}{0.95}

\newtcolorbox{fullpromptbox}{
  colback=lightgray,
  colframe=black,
  fonttitle=\bfseries,
  title=Logic Inference Prompt,
  enhanced
}

\usepackage{listings}
\newtcolorbox{myprocessbox}[1]{%
    enhanced,
    colback=cyan!5!white,
    colframe=cyan!75!black,
    fonttitle=\bfseries\large,
    title=#1,
    sharp corners,
    boxrule=0.8pt,
    parbox=false,
    before upper={\parindent0pt}%
}

\usepackage{tikz}
\usetikzlibrary{shapes.geometric, arrows.meta, positioning, decorations.pathreplacing, fit}

\usepackage{microtype}

\hypersetup{
    colorlinks=true,
    linkcolor=blue,
    filecolor=magenta,
    urlcolor=blue,
}

\begin{document}
\include{marcos}
\title{Large Language Model for OWL Proofs}
\author{Hui Yang}
\orcid{0000-0002-4262-4001}
\affiliation{
  \institution{The University of Manchester}
  \city{Manchester}
    \country{UK}
}
\email{hui.yang-2@manchetser.ac.uk}

\author{Jiaoyan Chen}
\orcid{https://orcid.org/0000-0003-4643-6750}
\affiliation{%
  \institution{The University of Manchester}
  \city{Manchester}
    \country{UK}
}
\email{jiaoyan.chen@manchester.ac.uk}

\author{Uli Sattler}
\orcid{https://orcid.org/0000-0003-4103-3389}
\affiliation{%
  \institution{The University of Manchester}
  \city{Manchester}
  \country{UK}
}
\email{uli.sattler@manchester.ac.uk}

\begin{abstract}
The ability of Large Language Models (LLMs) to perform reasoning tasks such as deduction has been widely investigated in recent years.
Yet, their capacity to generate proofs—faithful, human-readable explanations of why conclusions follow—remains largely underexplored. 
In this work, we study proof generation in the context of OWL ontologies, which are widely adopted for representing and reasoning over complex knowledge, by developing an automated dataset construction and evaluation framework.
Our evaluation encompassing three sequential tasks for complete proving: Extraction, Simplification, and Explanation, as well as an additional task of assessing Logic Completeness of the premise.
Through extensive experiments on widely used reasoning LLMs, we achieve important findings including: (1) Some models achieve overall strong results but remain limited on complex cases; (2) Logical complexity, rather than representation format (formal logic language versus natural language), is the dominant factor shaping LLM performance; and (3) Noise and incompleteness in input data substantially diminish LLMs' performance.
Together, these results underscore both the promise of LLMs for explanation with rigorous logics and the gap of supporting resilient reasoning under complex or imperfect conditions. Code and data are available at \url{https://github.com/HuiYang1997/LLMOwlR}.
\end{abstract}

\ccsdesc[500]{Computing methodologies~Description logics}
\ccsdesc[500]{Computing methodologies~Natural language generation}
\maketitle


\section{Introduction}

In recent years, analyzing the logical reasoning capabilities of Large Language Models (LLMs) has attracted considerable attention \cite{mondorf2024beyond,li2025system,liu2025logical}. Prior work has explored multiple facets of reasoning tasks, including deduction~\cite{DBLP:conf/ijcai/ClarkTR20,DBLP:conf/acl/TafjordDC21,DBLP:conf/iclr/Saparov023,he2023ontolama},  induction~\cite{DBLP:conf/emnlp/SinhaSDPH19}, constraint satisfaction problems~\cite{DBLP:journals/corr/abs-2502-01100} and combinations of multiple reasoning paradigms~\cite{DBLP:conf/iclr/YuJDF20,DBLP:conf/ijcai/LiuCLHWZ20,DBLP:journals/corr/abs-2505-16998,DBLP:conf/emnlp/TianLCX0J21,DBLP:conf/acl/ParmarPVN0MMB24,DBLP:conf/emnlp/HanS0QRZCPQBSWS24,xu2025large}. 
However, most studies focus narrowly on individual aspects of logical reasoning under idealized conditions, such as providing LLMs with complete premise sets and asking them to select correct answers from given options or verify reasoning processes.

The more comprehensive and realistic \textbf{proof construction and presentation} abilities of LLMs remain underexplored. Specifically, can LLMs effectively perform proof construction when given a conclusion and a set of potential premises? This task requires simultaneously identifying necessary logical information and generating correct, interpretable consequences—a significantly more challenging problem than traditional reasoning evaluation. Furthermore, how do LLMs perform when available premises are incomplete or only partially specified, reflecting real-world scenarios where perfect information is rarely available?

\begin{figure*}[t]
\centering
\scalebox{0.93}{
\begin{tikzpicture}[
    node distance=1.5cm,
    >=Stealth,
    arrow/.style={->, thick, >=Stealth},
    complexity/.style={font=\footnotesize, text=gray!70},
    input_box/.style={rectangle, rounded corners=8pt, fill=green!10, draw=green!60, thick, minimum width=7cm, minimum height=5cm, text width=7cm, font=\small},
    llm/.style={circle, fill=purple!20, draw=purple!60, thick, minimum size=1.5cm, text centered, font=\footnotesize},
    task_label/.style={font=\large\bfseries},
    combined_box1/.style={rectangle, rounded corners=5pt, fill=blue!5, draw=blue!60, thick, minimum width=8cm, minimum height=2cm, text width=7.8cm, font=\footnotesize},
    combined_box2/.style={rectangle, rounded corners=5pt, fill=orange!5, draw=orange!60, thick, minimum width=8cm, minimum height=2.2cm, text width=7.8cm, font=\footnotesize},
    combined_box3/.style={rectangle, rounded corners=5pt, fill=red!5, draw=red!60, thick, minimum width=9cm, minimum height=2.3cm, text width=8.5cm, font=\footnotesize}
    ]


\node[input_box, minimum height=1cm] (query_section) at (-6.2, 10) {
    
    \textbf{Query:} Why DomesticDog is subclass of CompanionAnimal?

    (i.e., prove conclusion: DomesticDog $\sqsubseteq$ CompanionAnimal)
    
    
};

\node[input_box, minimum height=4.5cm] (sources_section) at (-6.2, 5.5) {
    
    \textbf{ontology.org/animal:}\\
    DomesticDog $\sqsubseteq$ Mammal\\
    Mammal $\sqsubseteq$ Animal\\[0.1cm]
     \textbf{.\ .\ .}\\[0.1cm]
     \ 
    
    \textbf{petcare.com/taxonomy:}\\
    CompanionAnimal $\equiv$ Animal $\sqcap$ $\exists$hasOwner.Human\\
    Fish $\sqsubseteq$ $\exists$livesIn. AquaticEnvironment\\[0.1cm]
     \textbf{.\ .\ .}\\[0.1cm]
     \ 
    
    \textbf{biology.edu/species:}\\
    DomesticDog  $\equiv$ $\exists$hasOwner. Human $\sqcap$ Dog\\
    Bird $\sqsubseteq$ $\exists$canPerform. Flying\\[0.1cm]

    \textbf{.\ .\ .}\\[0.1cm]
     \

};
\node (llm) at (-1.3, 6.5) {\includegraphics[width=1.5cm]{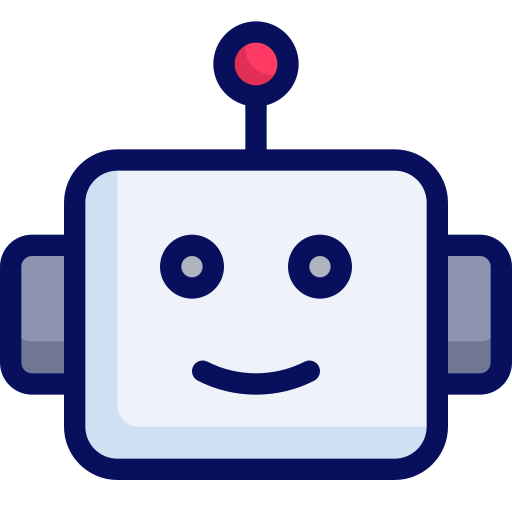}};

\node[task_label, blue!80] (task1_label) at (4, 10.9) {Task 1: Extraction};
\node[combined_box1] (task1_combined) at (4, 9.6) {
    \textbf{Extract relevant ontological axioms:}\\[0.2cm]
    
    • DomesticDog $\sqsubseteq$ Mammal (from ontology.org)\\
    • CompanionAnimal $\equiv$ Animal $\sqcap$ $\exists$hasOwner.Human (petcare.com)\\
    • DomesticDog  $\equiv$ $\exists$hasOwner. Human $\sqcap$ Dog (from biology.edu)\\[0.1cm]
};

\node[task_label, orange!80] (task2_label) at (4, 7.8) {Task 2: Simplification};
\node[combined_box2] (task2_combined) at (4, 6.5) {
     \textbf{Simplify axioms of complex ontological expressions}:\\[0.2cm]
    
    • CompanionAnimal $\equiv$ Animal $\sqcap$ $\exists$hasOwner.Human\\
    \quad → Animal $\sqcap$ $\exists$hasOwner.Human $\sqsubseteq$ CompanionAnimal\\
    • DomesticDog  $\equiv$ $\exists$hasOwner. Human $\sqcap$ Dog\\
    \quad → DomesticDog $\sqsubseteq$ $\exists$hasOwner.Human\\[0.1cm]
};

\node[task_label, red!80] (task3_label) at (4, 4.7) {Task 3: Explanation};
\node[combined_box3] (task3_combined) at (4, 3.3) {
    \textbf{Describe reasoning process}:\\[0.2cm]
    
    \textbf{Step 1:} DomesticDog $\sqsubseteq$ Animal $\sqcap$ $\exists$hasOwner.Human\\
    \quad \textbf{BECAUSE} DomesticDog $\sqsubseteq$ Mammal $\sqsubseteq$ Animal \textbf{AND}\\
    \quad DomesticDog $\sqsubseteq$ $\exists$hasOwner.Human\\[0.1cm]
    \textbf{Step 2:} DomesticDog $\sqsubseteq$ CompanionAnimal\\
    \quad \textbf{BECAUSE}  Animal $\sqcap$ $\exists$hasOwner.Human $\sqsubseteq$ CompanionAnimal  \textbf{AND} Step 1\\[0.1cm]
};


\draw[arrow, thick, purple!60] 
    (query_section.south) -- node[midway, right] {Relevant Web Ontologies} (sources_section.north);
\draw[arrow, thick, purple!60] 
  (sources_section.east |- llm.west) -- (llm.west);
\draw[arrow, thick, purple!60] (llm) |- (task1_combined.west);
\draw[arrow, thick, purple!60] (llm) -- (task2_combined.west);
\draw[arrow, thick, purple!60]  (llm) |- (task3_combined.west);

\draw[arrow, very thick, red!90] (8.8, 10.5) -- (8.8, 3);
\node[complexity, rotate=90] at (9.2, 6.5) {{\color{red}Increasing Difficulty}};

\end{tikzpicture}
}
\caption{LLMs for proof construction and presentation with Web ontologies}
\label{fig:framework}
\end{figure*}
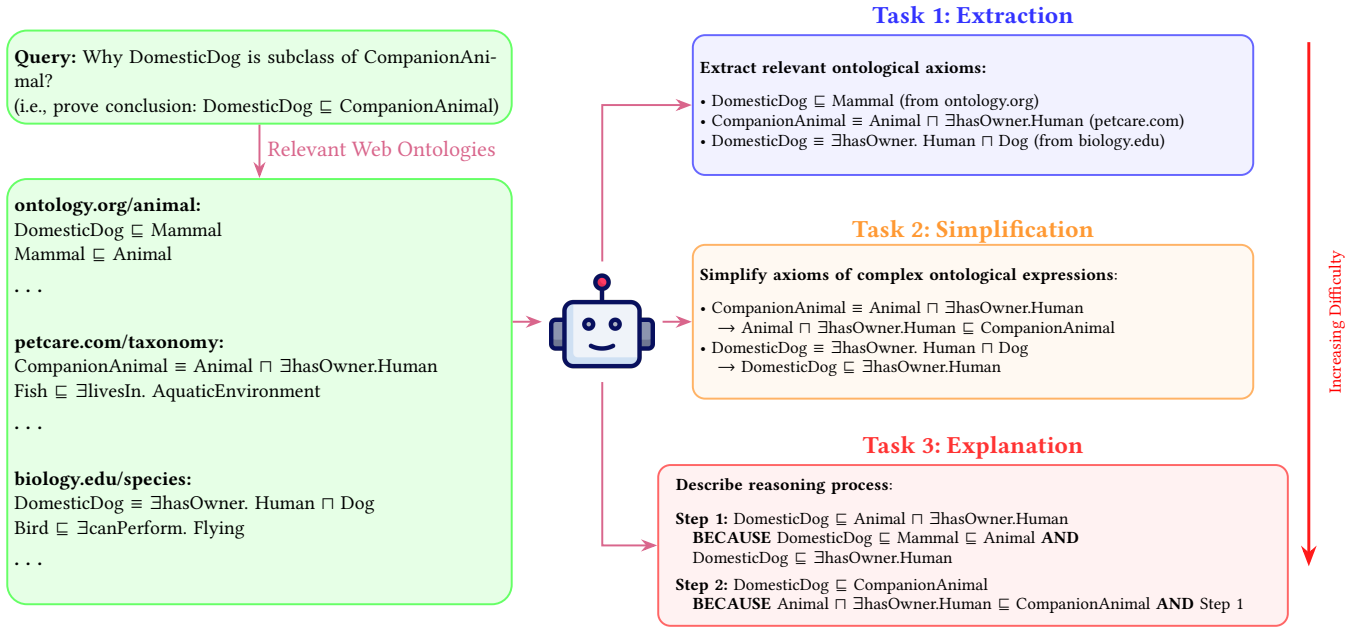

In this work, we investigate this realistic LLM evaluation problem on proof construction using OWL (Web Ontology Language) ontologies~\cite{DBLP:journals/ijinfoman/Fitz-GeraldW10a}, a widely adopted formalism for semantic reasoning across domains such as the Semantic Web~\cite{DBLP:journals/ker/Heder14}, healthcare~\cite{snomedct_international}, and finance~\cite{bennett2013financial}. OWL ontologies serve two key purposes: (i) they provide formal, specific, shared, and machine-readable knowledge representations as collections of \textbf{axioms}, and (ii) they enable automated reasoning over existing axioms to infer new knowledge, underpinned by Description Logic (DL).

Different from prior benchmarks that evaluate whether LLMs 
contain knowledge expressed in ontologies \cite{wu2023plms,bombierillms,OntoURL,mai2024llms},
or can perform deductive inference over ontologies~\cite{he2023ontolama,poulis2023reasoning,OntoURL}, we focus on evaluating the proving abilities of LLMs over OWL ontologies. To this end, we define three tasks of increasing difficulty (illustrated in Figure~\ref{fig:framework}):

\begin{enumerate}[leftmargin=*]
    \item \textbf{Extraction.} Identify and extract a set of axioms sufficient to infer the target conclusion. This evaluates both completeness (capturing all necessary supporting axioms) and conciseness (avoiding irrelevant axioms).
    \item \textbf{Simplification.} For each extracted supporting axiom, rewrite it into a simplified one by keeping only the essential parts that  contribute to the derivation of the target conclusion.
    \item \textbf{Explanation.} Given a conclusion and its supporting axioms, produce a logically valid, coherent, and understandable proof that demonstrates how the conclusion is inferred.
\end{enumerate}


Moreover, in real-world scenarios, logical information is often written informally in natural language and may be incomplete for inferring the target conclusion. To capture these scenarios, we introduce two additional evaluation dimensions: \textit{formal logic language versus natural language} and \textit{complete versus incomplete premises}. In the case of incomplete premises, a natural question arises: are the given premises sufficient to derive the target conclusion? We frame this as a \textbf{Logic Completeness} task within the incomplete-premise setting.

In summary, our evaluation goes beyond traditional reasoning tasks such as extracting relevant axioms and generating proofs. We also investigate more complex, realistic scenarios that classical ontology reasoners cannot address—specifically, cases where logical information is expressed informally in natural language, or where the set of premises is incomplete.

To facilitate systematic evaluation, we propose an automatic dataset construction and evaluation framework that can support arbitrary OWL ontologies of DL $\mathcal{EL}$, and investigate state-of-the-art reasoning LLMs of different sizes, including the DeepSeek-R1 family, the Qwen3 family, GPT-o4-mini and Magistral. Through extensive evaluation across diverse cases on three real-world ontologies, we gain a deeper understanding of LLMs’ reasoning capabilities with several important findings, including:

\begin{enumerate}[leftmargin=*]
\item  Some models, such as \textit{GPT-o4-mini} and \textit{Qwen3-32B}, achieve strong overall results but remain limited when confronted with conclusions that require complex derivation patterns beyond transitive closures. 
\item The primary determinant of performance lies in the underlying logical complexity of the task, rather than the form of expression (e.g., formal logic language vs. natural language).  
\item The greatest challenges arise from noise and incompleteness in the input: performance on reasoning tasks can decline considerably, dropping by up to 47\% with substantial irrelevant information and up to 38\% with missing premises.
\end{enumerate}
These findings provide valuable insights for future research on complex LLM reasoning problems such as explainable knowledge retrieval and generation, which aims to find exactly necessary materials from a knowledge base towards requests that require logical proving (such as answering natural language questions with attributions \cite{gao2023enabling,hu2025can}).

\section{Related Work}

\paragraph{LLM Evaluation on Logical Reasoning}
Evaluating the logical reasoning capabilities of LLMs has emerged as a central research direction. Prior work has primarily focused on tasks such as \textit{deductive reasoning}~\cite{DBLP:conf/ijcai/ClarkTR20,DBLP:conf/acl/TafjordDC21,DBLP:conf/iclr/Saparov023}, \textit{inductive reasoning}~\cite{DBLP:conf/emnlp/SinhaSDPH19}, \textit{constraint satisfaction problems}~\cite{DBLP:journals/corr/abs-2502-01100}, and hybrid settings that combine multiple reasoning paradigms (e.g., non-monotonic reasoning)~\cite{DBLP:conf/iclr/YuJDF20,DBLP:conf/ijcai/LiuCLHWZ20,DBLP:journals/corr/abs-2505-16998,DBLP:conf/emnlp/TianLCX0J21,DBLP:conf/acl/ParmarPVN0MMB24,DBLP:conf/emnlp/HanS0QRZCPQBSWS24}.
However, most existing benchmarks assume clean and complete premise sets. They are typically constructed by instantiating abstract rules from formal logics (e.g., propositional or first-order logic) with synthetic symbols or entities\footnote{E.g., given the premise in first-order logic $\{p(x),\ \forall x.\, p(x)\rightarrow q(x)\}$, one can derive $q(x)$. Replacing $x$ by a person (e.g., Tom) and $p, q$ by some attributes ($p$: likes apples; $q$: likes oranges), one obtain the natural language form with premise `\textit{Tom likes apples. Anyone who likes apples likes oranges}" and conclusion \textit{Tom likes oranges}".}. 
While systematic, such approaches abstract away from commonsense and domain-specific knowledge.
In contrast, our work leverages real-world ontologies from diverse domains. Ontologies encode both semantic richness and logical structure, making them a natural resource for evaluating LLMs’ ability to reason under incomplete premises while remaining grounded in real-world semantics. 

\paragraph{LLMs on Ontologies.}  
The application of LLMs to ontology engineering and reasoning has attracted increasing attention. Existing research can be broadly divided into two categories:  
(1) \textit{Ontology construction and validation}, which include constructing new ontologies axioms from text  \cite{lo2024endtoend,goyal2024silp_nlp,DBLP:conf/semweb/GiglouDA23,lippolis2025ontology}, evaluating ontology knowledge captured by pre-training LLMs ~\cite{mai2024llms,wu2023plms,bombierillms,OntoURL},  and validation of ontology correctness or abilities \cite{DBLP:journals/corr/abs-2507-14552,rebboud2024benchmarking,tufek2024validating} ; and  
(2) \textit{Subsumption prediction and inference}, which investigates whether LLMs can predict subsumptions in zero-shot, few-shot or finetuning settings and deductive reasoning capabilities~\cite{OntoURL,hit,ont,he2023ontolama,poulis2023reasoning}. 

However, these works devote less attention to the more complex proving context highlighted in the introduction. Such contexts involve explanation generation and logical evidence retrieval, both of which are crucial for practical applications.  
In this work, we address this gap by evaluating the proving ability of LLMs over ontologies, focusing on atomic subsumptions and their corresponding justifications~\cite{beacon,ignatievdebugging,kalyanpur2005debugging,DBLP:conf/ijcai/YangK0B23}.


\section{Preliminaries}\label{sec:pre_ont}

Ontologies represent knowledge using sets of statements, called \emph{axioms}, which describe concepts (unary predicates) and roles (binary predicates).  
In this work, we focus on $\mathcal{EL}$-ontologies, which are based on the Description Logic fragment of $\mathcal{EL}$ and are widely applied in domains such as biomedicine~\cite{DBLP:books/daglib/0041477}.  
Let $\NC = \{A,B,\ldots\}$, $\NR = \{r,t,\ldots\}$, and $\NI = \{a,b,\ldots\}$ denote pairwise disjoint sets of \emph{concept names}, \emph{role names}, and \emph{individual names}, respectively.  
\emph{$\mathcal{EL}$-concepts} are defined recursively as 
\(
C ::= \top \mid A \mid C \sqcap D \mid \exists r.C,
\) 
where $A \in \NC$, $r \in \NR$, and $C,D$ are $\mathcal{EL}$-concepts.  
An \emph{$\mathcal{EL}$-ontology} $\Omc$ is a finite set of axioms consisting of General Concept Inclusions (GCIs) axioms $C \sqsubseteq D$, where $C$ and $D$ are (complex) $\mathcal{EL}$-concepts, and facts about individuals in $\NI$. 
In this work, we focus mainly on the GCIs, by which complex logical relationships are usually defined.

\begin{example}
Consider a small university \emph{$\mathcal{EL}$-ontology} with atomic concepts $\texttt{Student}, \texttt{Graduate}, \texttt{Course}$, and role $\texttt{enroll}$. 
We can define two axioms: 
\(
\texttt{Graduate} \sqsubseteq \texttt{Student}, 
\texttt{Student} \sqsubseteq \exists \texttt{enroll.Course}.
\)
\end{example}
\noindent
From an ontology $\Omc$, new axioms can be inferred through logical reasoning. For example, given $A \sqsubseteq B$ and $B \sqsubseteq C$ in $\Omc$, one can derive $A \sqsubseteq C$, written as $\Omc \models A \sqsubseteq C$.  
For formal definition of entailment ``$\models$'' based on \emph{interpretations} and \emph{models}, please refer to~\cite{DBLP:books/daglib/0041477}.

\section{Dataset Construction}

Now, we describe the construction of our dataset, designed to evaluate the three tasks introduced earlier. The main idea is based on the concept of  \emph{justifications} for a conclusion, which captures the minimal set of axioms required to entail that conclusion. Intuitively, each justification provides the essential premise requirement for inferring the given entailment. The formal definition is given below:

\begin{defi}[Justification]
Let \(\mathcal{O}\) be an ontology and \(\alpha\) a conclusion with \(\mathcal{O} \models \alpha\). An axiom subset \(J \subseteq \mathcal{O}\) is a \emph{justification} for \(\alpha\) if \(J \models \alpha\), \textbf{and} no proper subset of \(J\) entails \(\alpha\) (i.e., no \(J' \subsetneq J\) such that \(J' \models \alpha\)).
\end{defi}

Justifications provide the foundation for the extraction, simplification, and explanation tasks in our dataset, as they capture the minimal set of axioms required to support each conclusion. It is important to note that a single conclusion may have multiple distinct justifications. For simplicity, in both the construction of the dataset and our experiments, we consider only the justification of minimal size, i.e., the one containing the fewest axioms.



The construction of the dataset proceeds in two main steps:
\begin{enumerate}[leftmargin=*] 
    \item We first randomly select subsumption axioms as the target conclusions along with their corresponding justifications as the golden answers for extraction. To ensure a well-distributed selection, we introduce a distance measure \emph{\subdist} for each subsumption and select axioms across different distance ranges (details in Section \ref{sec:sub_select}).
    
    \item We then introduce \emph{noisy axioms} based on semantic distances computed using embedding models. These noisy axioms do not belong to any justification for the given conclusion, which guarantees that the golden answer remains the only correct answer (details in Section \ref{sec:noise}).
\end{enumerate}

During evaluation, we shuffle the order of both the original and noisy axioms, and prompt the LLMs to generate detailed explanations using prompt learning techniques (for details see Appendix  \ref{app:implementation}).

\subsection{Selecting Conclusions}\label{sec:sub_select}

Given an ontology $\Omc$, our construction method focus on inferred atomic subsumptions of the form $\Omc \models A \sqsubseteq B$, where $A$ and $B$ are atomic concepts (i.e., $A,B\in \texttt{N}_C$). To ensure that the selected subsumptions cover broad cases, we introduce a  metric over atomic subsumptions, denoted as \textbf{\subdist}, and select them  uniformly across different distance levels. The \subdist serves as a heuristic for estimating the ``reasoning length'' of an atomic subsumption, measured by the number of intermediate atomic concepts involved. Specifically:

\begin{enumerate}[leftmargin=*] 
    \item \textbf{Distance 1 (Direct Subsumption):} A subsumption \(A \sqsubseteq B\) has distance 1 if \(\Omc \models A \sqsubseteq B\) and there exists no atomic concept \(A' \notin \{A, B\}\) such that both \(\Omc \models A \sqsubseteq A'\) and \(\Omc \models A' \sqsubseteq B\).
    
    \item \textbf{Distance \(k\) (Indirect Subsumption for $k \geq 2$):} More generally, for an atomic subsumption $\Omc \models A' \sqsubseteq B'$, the \subdist of $A' \sqsubseteq B'$ is defined as the minimal number $k$ such that there exists a chain of direct subsumptions 
    \[
        A_0 \sqsubseteq A_1, \; A_1 \sqsubseteq A_2, \; \ldots, \; A_{k-1} \sqsubseteq A_k, \text{ where } A_0 = A', A_k = B'.
    \]
\end{enumerate}

\begin{rem}\label{rem:subdits}
It is worth noting that a subsumption \(\Omc \models A \sqsubseteq B\) with a higher \subdist\ typically has justifications involving more axioms. When the \subdist\ coincides with the number of justifications, the derivation pattern often follows directly from transitive closure. For example, in
\(
\Omc = \{A \sqsubseteq A_1,\; A_1 \sqsubseteq A_2,\; \ldots,\; A_{k-1} \sqsubseteq A_k\},
\)
the subsumption \(\Omc \models A \sqsubseteq A_k\) has both \subdist\ and number of justifications equal to \(k\), corresponding to the straightforward transitive chain
\(A \sqsubseteq A_1 \sqsubseteq \ldots \sqsubseteq A_k\).

In contrast, when the justifications contain more axioms than the \subdist, the derivation usually involves more intricate reasoning patterns, and we classify such cases as \emph{complex}. For instance, consider
\(
\Omc' = \{A \sqsubseteq \exists r.B,\; B \sqsubseteq B_1,\; \exists r.B_1 \sqsubseteq A_1\}.
\)
Here, the subsumption \(\Omc' \models A \sqsubseteq A_1\) has \subdist\ 1, since there is no intermediate concept \(A'\) with \(\Omc' \models A \sqsubseteq A' \sqsubseteq A_1\). However, the number of justifications is 3, reflecting the use of more sophisticated reasoning based on the existential operator ``\(\exists r\)''.
\end{rem}






\subsection{Selecting Noisy Axioms}\label{sec:noise}
Since real-world ontologies can be extremely large (e.g., Snomed CT contains over 350,000 axioms) and it is impractical and costly to feed them all into LLMs directly for evaluating the extraction task, we construct the input by adding noisy axioms into the golden justification as follows.  
We introduce noisy axioms primarily based on their semantic interpretations. Specifically, we first translate both the given conclusion and each axiom in the ontology into natural language sentences according to their semantic meanings. Then, we select noisy axioms by measuring their semantic distance using sentence embedding models, which evaluate the degree of semantic similarity between sentences.

In details, following~\cite{he2023ontolama}, for an axiom of the form $C \sqsubseteq D$ (resp. $C \equiv D$), the corresponding sentence is:  
$\mathcal{V}(C)\textit{ is a subclass of } \mathcal{V}(D)$ 
(resp.  $\mathcal{V}(C)\textit{ is equivalent to } \mathcal{V}(D)$),
where $\mathcal{V}(C)$ and $\mathcal{V}(D)$ denote the verbalization of the concepts $C$ and $D$, generated compositionally by the following rules:  
\begin{align*}
\mathcal{V}(C \sqcap D) &= "\mathcal{V}(C) \text{ and } \mathcal{V}(D)", \\
\mathcal{V}(\exists r.C) &= "\text{something that } \mathcal{V}(r) \text{ some } \mathcal{V}(C)".    
\end{align*}
For instance, we have  
\(\mathcal{V}(\text{Person} \sqcap \text{Student}) = \)``person and student", 
\(\mathcal{V}(\exists \text{isParentOf}.\text{Person}) = \)``something that is parent of some person". 
Similarly, conclusions $C\sqsubseteq D$ are transformed into query-like sentences of the form:  
\(
 \textit{``Why is } \mathcal{V}(C)\textit{ a subclass of } \mathcal{V}(D)\textit{?"}. 
\)
Finally, noisy axioms are selected based on their embedding distances, with embedding vectors computed using the sentence embedding model BGE~\cite{DBLP:journals/corr/abs-2309-07597}.

One could consider extracting noisy axioms based on logical structures, such as the hypergraph-based representations proposed by \cite{DBLP:conf/cade/YangMB22,DBLP:conf/ecai/Ludwig014} or ontology modules~\cite{konev2008semantic,gatens2014lower,sattler2009kind,Hui,DBLP:conf/ijcai/YangK0B23,zoomingin}. However, this approach requires a more detailed discussion of the selection process and additional effort to control the size of the selected result. For simplicity, we instead rely on semantic meaning, which is more convenient and better suited to our experimental setup. In particular, our experiments also investigate the impact of the natural language forms of axioms (Section~\ref{sec:nature_language}), rather than focusing solely on formal logical expressions.

\begin{figure}
    \centering
    \includegraphics[width=0.85\linewidth]{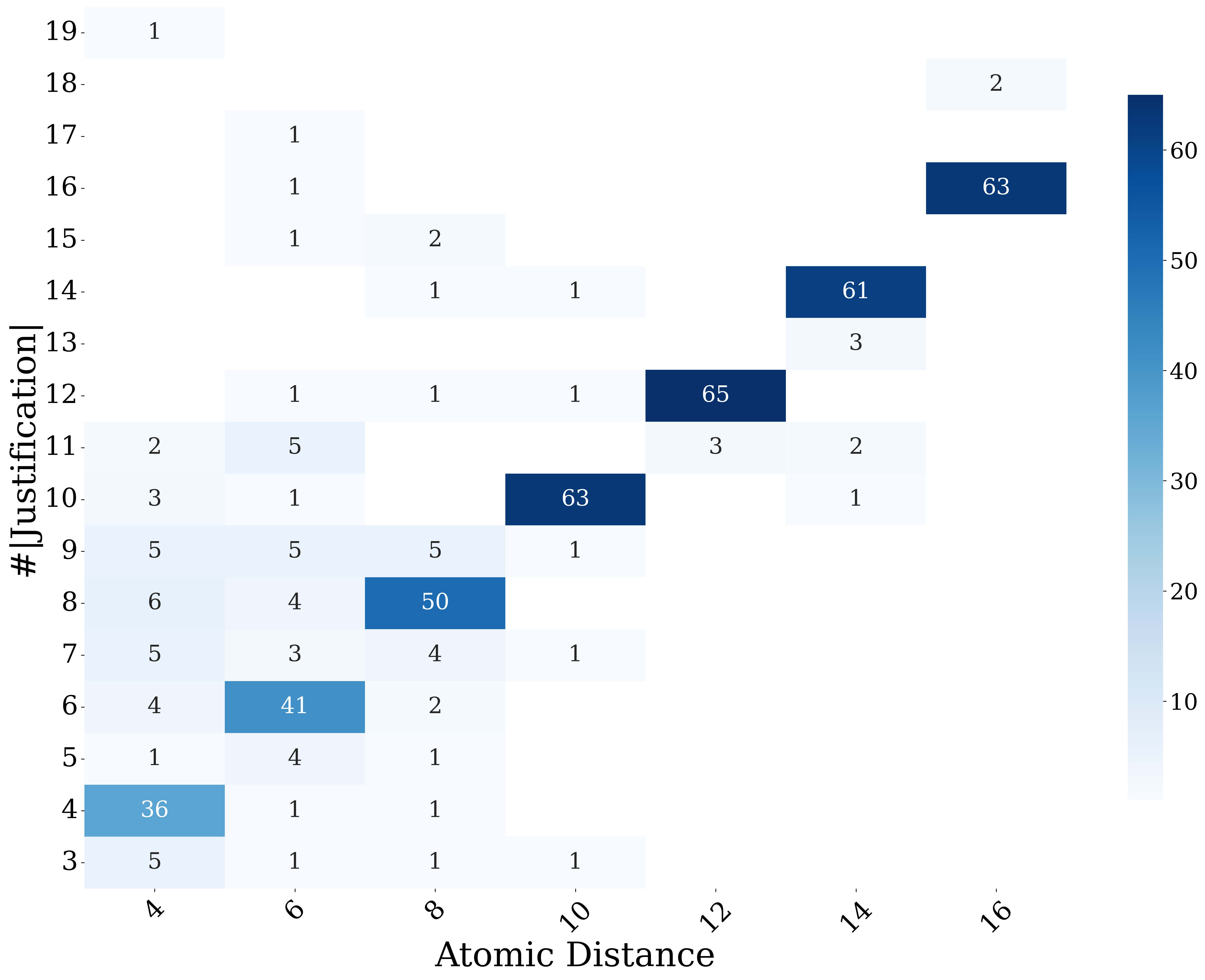}
    \vspace{-0.35cm}
    \caption{Distribution of the selected conclusions (subsumptions).}
    \label{fig:stat_selected}
\end{figure}

\subsection{Constructed Dataset}
We use three widely used, large-scale ontologies—\textbf{Snomed CT}~\cite{snomedct_international}, \textbf{Foodon}~\cite{DBLP:conf/icbo/GriffithsDBHBH16}, and \textbf{GO-Plus}~\cite{ashburner2000gene}, which have been used for evaluation in many previous works \cite{chen2025ontology}. 
They may infer millions of subsumptions (detailed statistic given by Figure \ref{fig:stat_subdist} in Appendix). 
Due to the computer source limitations, we only select part of them  in our evaluation. It should be noted that other researchers can use our code for generating more sample according to their interest. 
Specifically, from Snomed CT, we randomly extract 350 conclusions. For the other two ontologies, whose logical structures are simpler, we select 70 conclusions from each. All conclusions are chosen with an even \subdist range uniformly spanning from 4 to 16. We filter out the non-$\mathcal{EL}$ axioms that contains non-EL constructors (e.g., $\neg$).  The distribution of these selected conclusions (subsumptions), over different atomic-distances and different justification sizes (i.e., numbers of axioms) are shown in the Figure \ref{fig:stat_selected}.

\section{Evaluation Methods and Settings}

\subsection{Methods}  
We test the performance based on prompt-based method with two options: (1) supplying a set of inference rules to guide reasoning over $\mathcal{EL}$-ontologies, as illustrated in Figure~\ref{fig:inf_rule}; or (2) providing examples, such as the one in Figure~\ref{fig:prompt-example} (detailed in Appendix \ref{app:implementation}).



The evaluated language models span both open-source and commercial systems: \textbf{(1) Open-source models:} The 32B models comprise \textit{DeepSeek-R1-Distill-Qwen-32B}, \textit{Qwen3-32B}, and \textit{Qwen2.5-32B}, the 24B model by \textit{Magistral-Small-2506}, and the 8B models \textit{Qwen3-8B}, \textit{DeepSeek-R1-Distill-Qwen-8B}, and \textit{DeepSeek-R1-Distill-Llama-8B}; 
\textbf{(2) Commercial baseline:} \textit{GPT-o4-mini}, a cost-effective commercial model that demonstrates strong performance. 
This organization enables systematic comparisons across model scale, architecture, and source type. For all LLMs tested, the max token limit is 5,000, with temperature fixed at 0 as convention to ensure reproducibility and enable fair comparison across models. 

\begin{figure}
    \centering
    \small
    \begin{tcolorbox}[title=Inference Rules,colback=lightgray,colframe=black]
    \texttt{\#\# Inference Rules}
    
    \texttt{1. Subsumption: If A }$\sqsubseteq$\texttt{ B and B }$\sqsubseteq$\texttt{ C, then A }$\sqsubseteq$\texttt{ C}
    
    \texttt{2. Conjunction: If A }$\sqsubseteq$\texttt{ B, A }$\sqsubseteq$\texttt{ C, then A }$\sqsubseteq$\texttt{ B }$\sqcap$\texttt{ C}
    
    \texttt{3. Existential: If A }$\sqsubseteq$ $\exists \texttt{r.C}$\texttt{ and C }$\sqsubseteq$\texttt{ D, then A }$\sqsubseteq$ $\exists \texttt{r.D}$
    
    \texttt{4. Equivalence: If A }$\equiv$ \texttt{ B, then A }$\sqsubseteq$\texttt{ B and B }$\sqsubseteq$\texttt{ A}
    
    \texttt{5. Conjunction Inference: If A }$\sqsubseteq$\texttt{ B }$\sqcap$\texttt{ C, then A }$\sqsubseteq$\texttt{ B and A }$\sqsubseteq$\texttt{ C}
    
    \vspace{0.5em}
    
    \texttt{Here, A, B, C, ... means concept, r means roles. }$\sqsubseteq$\texttt{ means subclass, }$\equiv$\texttt{ means equivalence, }$\exists r.A$\texttt{ means the set of elements in A that have a relationship r with A.}
    
    \end{tcolorbox}
    \vspace{-0.35cm}
    \caption{Inference Rules (for $\mathcal{EL}$-ontologies)}
    \label{fig:inf_rule}
\end{figure}

\begin{figure}
    \centering
    \small
    \begin{tcolorbox}[title=Example,colback=lightgray,colframe=black]
\texttt{\#\# Example}

\texttt{Given the following axioms:} \\
\texttt{[1] A }$\equiv$ $\exists \texttt{r.B}$ \\
\texttt{[2] C }$\sqsubseteq$\texttt{ B }$\sqcap$\texttt{ H }$\sqcap$\texttt{ I} \\
\texttt{[3] D }$\equiv$$\exists \texttt{r.C}$ $\sqcap$\texttt{ G} \\
\texttt{[4] E }$\equiv$\texttt{ A }$\sqcap$\texttt{ F} \\
\texttt{[5] D }$\sqsubseteq$\texttt{ F }$\sqcap$\texttt{ J}\\
\texttt{[6] D }$\sqsubseteq$\texttt{ K}

\vspace{0.3em}
\texttt{The desired explanation for deriving the conclusion D }$\sqsubseteq$\texttt{ E is as follows:}

\vspace{0.3em}
\texttt{**AXIOMS\_USED**: 1,2,3,4,5}

\vspace{0.1em}
\texttt{**SIMPLIFY**:} \\
\texttt{[1] A }$\equiv$ $\exists \texttt{r.B}$ \texttt{ → }$\exists \texttt{r.B}$ $\sqsubseteq$\texttt{ A} \\
\texttt{[2] C }$\sqsubseteq$\texttt{ B }$\sqcap$\texttt{ H }$\sqcap$\texttt{ I } \texttt{→ C }$\sqsubseteq$\texttt{ B} \\
\texttt{[3] D }$\equiv$ $\exists \texttt{r.C}$ $\sqcap$\texttt{ G \texttt{→ D }$\sqsubseteq$ $\exists \texttt{r.C}$} \\
\texttt{[4] E }$\equiv$\texttt{ A }$\sqcap$\texttt{ F \texttt{→ A }$\sqcap$\texttt{ F }$\sqsubseteq$\texttt{ E}} \\
\texttt{[5] D }$\sqsubseteq$\texttt{ F }$\sqcap$\texttt{ J \texttt{→ D }$\sqsubseteq$\texttt{ F}}

\vspace{0.1em}
\texttt{**DERIVE**:} \\
\texttt{STEP1: [1,2,3] }$\vdash$\texttt{ D }$\sqsubseteq$\texttt{ A} \\
\texttt{    EXPLANATION: D }$\sqsubseteq$ $\exists \texttt{r.C}$\texttt{ $\wedge$ C }$\sqsubseteq$\texttt{ B }$\Rightarrow$\texttt{ D }$\sqsubseteq$ $\exists \texttt{r.B} \wedge \exists \texttt{r.B}$ $\sqsubseteq$\texttt{ A }$\Rightarrow$\texttt{ D }$\sqsubseteq$\texttt{ A}

\vspace{0.2em}
\texttt{STEP2: [STEP1,4,5] }$\vdash$\texttt{ D }$\sqsubseteq$\texttt{ E} \\
\texttt{    EXPLANATION: D }$\sqsubseteq$\texttt{ A $\wedge$ D }$\sqsubseteq$\texttt{ F }$\Rightarrow$\texttt{ D }$\sqsubseteq$\texttt{ A }$\sqcap$\texttt{ F $\wedge$ A }$\sqcap$\texttt{ F }$\sqsubseteq$\texttt{ E }$\Rightarrow$\texttt{ D }$\sqsubseteq$\texttt{ E}
\end{tcolorbox}
\vspace{-0.35cm}
\caption{An example of the given input and desired output. The output is supposed to consist of three parts, \texttt{AXIOMS\_USED}, \texttt{SIMPLIFY}, and \texttt{DERIVE}, correspond to the tasks \textit{Extraction}, \textit{Simplification}, and \textit{Derivation}, respectively.}
    \label{fig:prompt-example}
\end{figure}

\subsection{Metrics} 

\textbf{Format Correctness Rate}
This metric measures the percentage of samples in which the LLM produces outputs that adhere to the required format. That is, the output includes the three sections \texttt{AXIOMS\_USED}, \texttt{SIMPLIFY}, and \texttt{DERIVE}, as illustrated in Figure \ref{fig:prompt-example}.
Note that a high generation success rate indicates that the reasoning LLMs have robust formatting ability or their reasoning processes are relatively brief, 
but it does not guarantee the correctness of the results.

\noindent
\textbf{Extraction}
We evaluate extraction performance using the Jaccard Similarity, defined as (\( S \) denotes the output axiom set in the section \texttt{AXIOM\_USED} and \( S_{\text{gt}} \) denotes the ground-truth set):
\[
\text{Jaccard Similarity} = \frac{|S \cap S_{\text{gt}}|}{|S \cup S_{\text{gt}}|}
\]

\noindent
\textbf{Simplification}
Performance on the simplification task is evaluated using the following metrics, which assess both logical correctness and efficiency:

\begin{enumerate}[leftmargin=*]
    \item \textit{Axiom-wise Accuracy:} A single-axiom simplification is correct if its simplified form is logically entailed by the original axiom (e.g., \( \{A \equiv C\} \models C \sqsubseteq A \)).

    \item \textit{Overall Accuracy:} A simplification is fully correct if all individual axioms are correctly simplified \textbf{and} the target conclusion remains derivable from the simplified forms.
    
      \item \textit{Length-drop Percentage:} This metric quantifies the efficiency of simplification by measuring the total reduction in axiom length\footnote{Length is computed by assigning weight 1 to each logical operator, role, and concept, and weight 2 to equivalence $(\equiv)$.}.

\end{enumerate}

\noindent
\textbf{Derivation}
Analogous to the simplification case, we measure performance on the derivation task using:

\begin{enumerate}[leftmargin=*]
    \item \textit{Step-wise Accuracy:} A derivation step is correct if the conclusion logically follows from its premises \textbf{and} all premises are either given axioms or results of previously validated steps.

    \item \textit{Overall Accuracy:} A derivation is fully correct if all individual steps meet the step-level criteria \textbf{and} the target conclusion is successfully derived.

    \item \textit{Derivation Steps:} This measures the total steps that the output derivations have.
\end{enumerate}

\begin{rem}\label{rem:weighted_value}
Most of the metrics introduced above can only be computed when the output is in the correct format. 
To account for outputs with incorrect formatting, we also use a \textbf{weighted} version of each metric, obtained by multiplying the format-correctness rate with the corresponding metric.
\end{rem}

\begin{table*}[htbp]
\centering
\caption{Comparison of model performance (results presented as performance  with/withthout inference rules). \textbf{Format-Correct} denotes Format Correctness  Rate of the generated response. The top 3 results for each metric are highlighted in \textcolor{red}{red} (1st), \textcolor{blue}{blue} (2nd), and \textcolor{orange}{orange} (3rd).}
\vspace{-0.25cm}
\label{tab:main_result}
\resizebox{\textwidth}{!}{
\begin{tabular}{lcccccccc}
\toprule
\multirow{2}{*}{Model} 
& \textbf{Format-} 
& \textbf{Jaccard}
& \multicolumn{3}{c}{\textbf{Simplification Acc.}} 
& \multicolumn{3}{c}{\textbf{Derivation Acc.}}  \\
\cmidrule(lr){4-6} \cmidrule(lr){7-9}
& \textbf{Correct}&  \textbf{ Avg.} & \textbf{axiom-wise} & \textbf{overall} &  \textbf{length-drop} & \textbf{step-wise} & \textbf{overall} & \textbf{\#|Steps|}  \\
\midrule
\multicolumn{8}{c}{\textbf{Snomed CT}} \\
\midrule
GPT-o4-mini & 83.43 / 92.26 & {\color{blue}99.19} / {\color{red}99.66} & {\color{red}98.66} / {\color{orange}97.97} & {\color{blue}91.78} / {\color{red}91.93} & 27.04 / 26.55 & {\color{red}97.21} / {\color{orange}93.06} & {\color{red}93.49} / {\color{blue}86.02} & 7.86 / 7.02 \\
DeepSeek\_R1\_Llama-8B & 58.86 / 60.86 & 68.18 / 66.21 & 78.57 / 76.14 & 34.58 / 29.78 & 23.85 / 24.90 & 49.39 / 51.48 & 14.02 / 14.22 & {\color{blue}4.59} / {\color{red}4.20} \\
DeepSeek\_R1\_Qwen-32B & {\color{orange}97.43} / 86.86 & 42.23 / 35.63 & 96.55 / 93.25 & 8.00 / 6.02 & 33.29 / {\color{red}40.21} & 53.39 / 58.77 & 7.71 / 7.16 & {\color{orange}4.89} / 4.95 \\
DeepSeek\_R1\_Qwen-8B & 15.71 / 16.57 & 52.63 / 49.34 & 77.16 / 72.96 & 40.48 / 29.41 & 15.82 / 23.91 & 63.85 / 73.44 & 8.33 / 7.84 & 8.17 / 6.90 \\
Qwen2.5-32B & {\color{blue}99.71} / {\color{red}100.00} & 39.81 / 39.41 & 93.26 / 92.37 & 5.71 / 6.29 & 20.33 / 25.57 & 60.97 / 58.80 & 6.57 / 7.14 & 7.07 / 5.68 \\
Qwen3-32B & 73.71 / 29.71 & {\color{orange}95.87} / 82.28 & {\color{blue}97.97} / 97.12 & {\color{orange}69.47} / 51.33 & 30.24 / 29.70 & {\color{blue}94.15} / 88.03 & {\color{orange}72.14} / 61.95 & 7.31 / 6.50 \\
Qwen3-8B & 16.57 / 22.29 & 86.87 / 86.78 & 94.32 / 94.72 & 67.19 / 66.67 & 25.92 / 31.64 & 88.71 / 88.64 & 51.56 / 50.62 & 9.83 / 7.17 \\
Magistral-Small-2506 & 33.71 / 29.14 & 84.77 / 81.55 & 87.62 / 89.29 & 38.56 / 36.96 & {\color{orange}35.36} / {\color{blue}36.49} & 78.55 / 88.90 & 32.68 / 34.78 & 6.22 / 6.07 \\
\midrule
\multicolumn{8}{c}{\textbf{GO-Plus}} \\
\midrule
GPT-o4-mini & {\color{red}100.00} / {\color{blue}100.00} & {\color{red}100.00} / {\color{blue}100.00} & {\color{red}100.00} / 98.65 & {\color{red}98.11} / {\color{blue}98.11} & 2.24 / 3.23 & 98.47 / {\color{orange}98.67} & {\color{red}98.11} / {\color{blue}94.34} & 6.15 / 5.66 \\
DeepSeek\_R1\_Llama-8B & 79.25 / 66.04 & 84.73 / 69.71 & 89.21 / 90.59 & 69.05 / 64.86 & 2.90 / 4.39 & 40.00 / 41.77 & 26.19 / 8.11 & {\color{blue}3.93} / 4.27 \\
DeepSeek\_R1\_Qwen-32B & {\color{orange}98.11} / 81.13 & 47.18 / 36.69 & 97.04 / 77.87 & 13.46 / 1.96 & 1.67 / {\color{orange}7.76} & 56.67 / 52.07 & 17.31 / 15.69 & {\color{orange}4.04} / {\color{red}3.31} \\
DeepSeek\_R1\_Qwen-8B & 52.83 / 49.06 & 54.42 / 61.90 & 81.51 / 93.29 & 50.00 / 68.75 & {\color{red}12.57} / 7.23 & 62.48 / 76.40 & 11.76 / 31.25 & 14.97 / 12.31 \\
Qwen2.5-32B & 96.23 / 96.23 & 45.44 / 46.94 & 97.51 / 95.35 & 9.80 / 5.88 & 0.09 / 3.02 & 54.73 / 59.57 & 11.76 / 15.69 & 6.63 / 4.61 \\
Qwen3-32B & 73.58 / 35.85 & {\color{orange}96.73} / 86.39 & {\color{orange}99.48} / 98.34 & {\color{orange}95.00} / 80.00 & 4.00 / 3.03 & {\color{red}99.16} / 97.62 & {\color{orange}92.50} / 60.00 & 5.97 / 4.20 \\
Qwen3-8B & 32.08 / 52.83 & 85.45 / 95.92 & 98.21 / {\color{blue}100.00} & 84.21 / 82.76 & 1.42 / 0.80 & 94.78 / {\color{blue}98.80} & 78.95 / 82.76 & 7.05 / 8.59 \\
Magistral-Small-2506 & 54.72 / 54.72 & 84.08 / 84.00 & 92.23 / 92.79 & 55.88 / 65.62 & {\color{blue}10.30} / 6.78 & 73.62 / 84.06 & 44.12 / 37.50 & 6.91 / 6.47 \\
\midrule
\multicolumn{8}{c}{\textbf{Foodon}} \\
\midrule
GPT-o4-mini & {\color{red}100.00} / {\color{blue}100.00} & {\color{red}100.00} / {\color{blue}100.00} & {\color{red}99.29} / {\color{blue}99.18} & {\color{red}100.00} / {\color{blue}97.14} & 1.91 / 3.55 & {\color{blue}95.07} / {\color{red}95.09} & {\color{orange}88.57} / {\color{red}91.43} & 6.96 / 6.40 \\
DeepSeek\_R1\_Llama-8B & 70.00 / 64.29 & 87.62 / 76.31 & 91.07 / 83.45 & 56.00 / 54.35 & 1.16 / 1.15 & 55.51 / 52.63 & 26.00 / 15.22 & 5.08 / {\color{blue}4.13} \\
DeepSeek\_R1\_Qwen-32B & {\color{orange}95.71} / 92.86 & 43.41 / 38.56 & {\color{orange}97.67} / 81.18 & 15.94 / 11.43 & 2.87 / 6.58 & 72.73 / 58.27 & 11.59 / 15.71 & 5.58 / {\color{red}3.80} \\
DeepSeek\_R1\_Qwen-8B & 48.57 / 32.86 & 42.08 / 58.57 & 70.70 / 90.74 & 50.00 / 46.67 & {\color{red}12.31} / {\color{blue}10.81} & 67.30 / 86.40 & 5.26 / 30.00 & 20.76 / 9.07 \\
Qwen2.5-32B & 94.29 / 94.29 & 39.95 / 39.08 & 96.65 / 95.06 & 12.12 / 12.12 & 3.09 / 2.03 & 62.66 / 56.62 & 21.21 / 15.15 & 6.05 / {\color{orange}4.58} \\
Qwen3-32B & 74.29 / 37.14 & {\color{orange}97.58} / 90.11 & 97.65 / 95.57 & 90.38 / 73.08 & 5.29 / 6.91 & 91.36 / {\color{orange}91.39} & {\color{blue}90.38} / 65.38 & 7.79 / 5.81 \\
Qwen3-8B & 28.57 / 50.00 & 89.52 / 97.20 & 87.81 / 96.85 & 85.71 / {\color{orange}91.43} & 7.28 / 4.92 & 88.33 / 84.23 & 66.67 / 77.14 & 5.71 / 10.14 \\
Magistral-Small-2506 & 51.29 / 44.99 & 87.64 / 83.85 & 87.95 / 93.59 & 58.79 / 61.54 & {\color{orange}9.36} / 6.48 & 78.52 / 79.38 & 53.27 / 51.65 & 5.85 / 6.53 \\
\bottomrule
\end{tabular}}
\end{table*}

\section{Evaluation Results}
We begin with a standard setting, where the input axioms are expressed in formal logic and contain the complete justification of the output conclusion as well as the same number of noisy axioms as the justification. In Section~\ref{sec:complex_case}, we extend our analysis to the more complex settings in which the input contains more noisy axioms, axioms expressed in natural language, or incomplete premises.

\subsection{Standard Setting}



\subsubsection{Performance of Different Evaluation Methods}  
The main results across all datasets are summarized in Table~\ref{tab:main_result}. We have the following observations for the overall best performance.
For the \textbf{format correctness rate}, we observe that all relatively large models ($\geq$32B) achieve strong performance, typically above 70\% when inference rules are provided.  
Compared to GO-Plus and Foodon, the format correctness on Snomed CT is generally lower, which may be attributed to the comparatively higher complexity of derivations in Snomed CT. This observation is also supported by the relatively low accuracy observed on Snomed CT in the extraction, simplification, and derivation tasks discussed below.

For the \textbf{extraction} task, \textit{GPT-o4-mini} and \textit{Qwen3-32B} achieve the best performance, with Jaccard similarity scores consistently exceeding 95\% across different ontologies. In contrast, some models such as \textit{Qwen2.5-32B} and \textit{DeepSeek-R1-Qwen-32B} demonstrate high format correctness but very low Jaccard similarity (below 50\%). This discrepancy suggests that these models primarily capture structural patterns rather than genuinely understanding the reasoning process.  

For the \textbf{simplification} task, we find that \textit{GPT-o4-mini} consistently achieves the highest accuracy, exceeding 90\% across all the three datasets. \textit{Qwen3-32B} ranks second, performing substantially better than the remaining models but still lagging by around 40\% behind \textit{GPT-o4-mini} in weighted overall accuracy on Snomed CT. However, the performance gap is much smaller ($<10\%$) on other two datasets: GO-Plus and Foodon. 

Moreover, when evaluating simplification efficiency based on the length-drop percentage, we find that \textit{Snomed CT} exhibits substantially higher reductions (typically $>$25\%) compared to \textit{GO-Plus} and \textit{Foodon} (typically $<$5\%). This is primarily because axioms in Snomed CT are generally more complex than those in GO-Plus and Foodon, which usually consist of only atomic subsumptions (i.e., $A \sqsubseteq B$) that cannot be further simplified.

The performance on \textbf{derivation} tasks follows a similar pattern to simplification: \textit{GPT-o4-mini} achieves the best results, followed by \textit{Qwen3-32B}, which lags up to 33\% behind \textit{GPT-o4-mini} in weighted overall accuracy. 
    Additionally, we find that derivations typically require 7--8 steps on average. However, some models deviate substantially (e.g., $>$20 steps for \textit{DeepSeek-R1-Qwen-8B} on Foodon), which is not reliable given their low overall derivation accuracy.

\begin{figure}
    \centering
    \begin{subfigure}{\linewidth}
        \centering
        \includegraphics[width=0.85\linewidth]{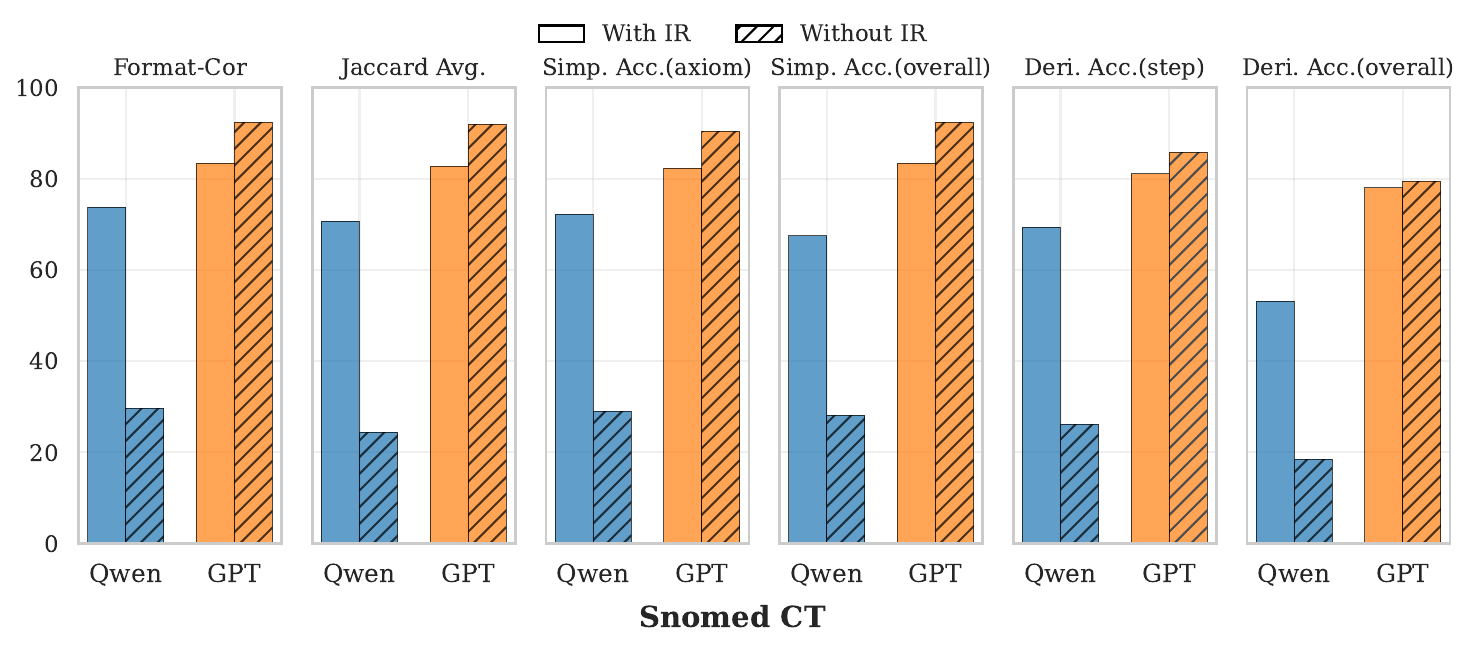}
    \end{subfigure}  
    \begin{subfigure}{\linewidth}
        \centering
        \includegraphics[width=0.85\linewidth]{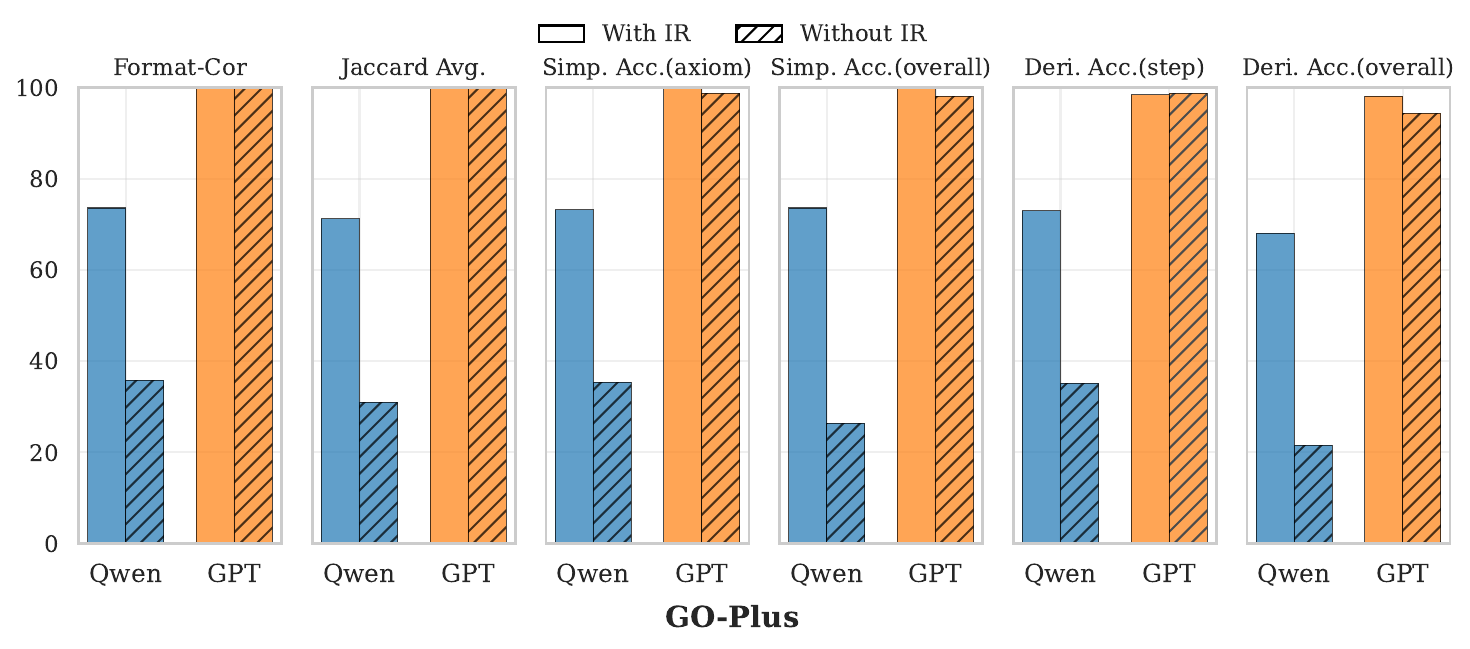}
    \end{subfigure}
    \begin{subfigure}{\linewidth}
        \centering
        \includegraphics[width=0.85\linewidth]{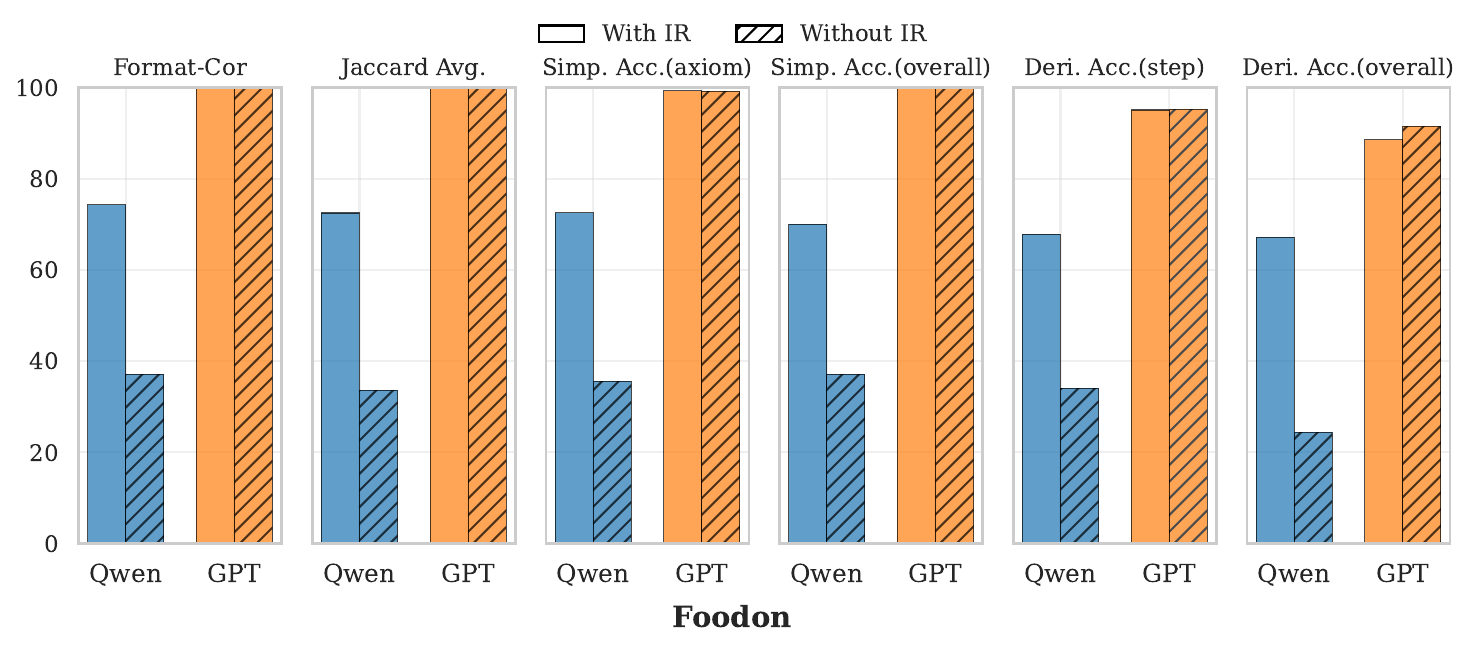}
    \end{subfigure}
    \caption{Comparison (weighted) of \textit{Qwen3-32B} and \textit{GPT-o4-mini} with/without Inference Rules (IR).}
    \label{fig:impact_ir}
\end{figure}

\begin{figure}
    \centering
    \begin{subfigure}{\linewidth}
        \centering
        \includegraphics[width=0.85\linewidth]{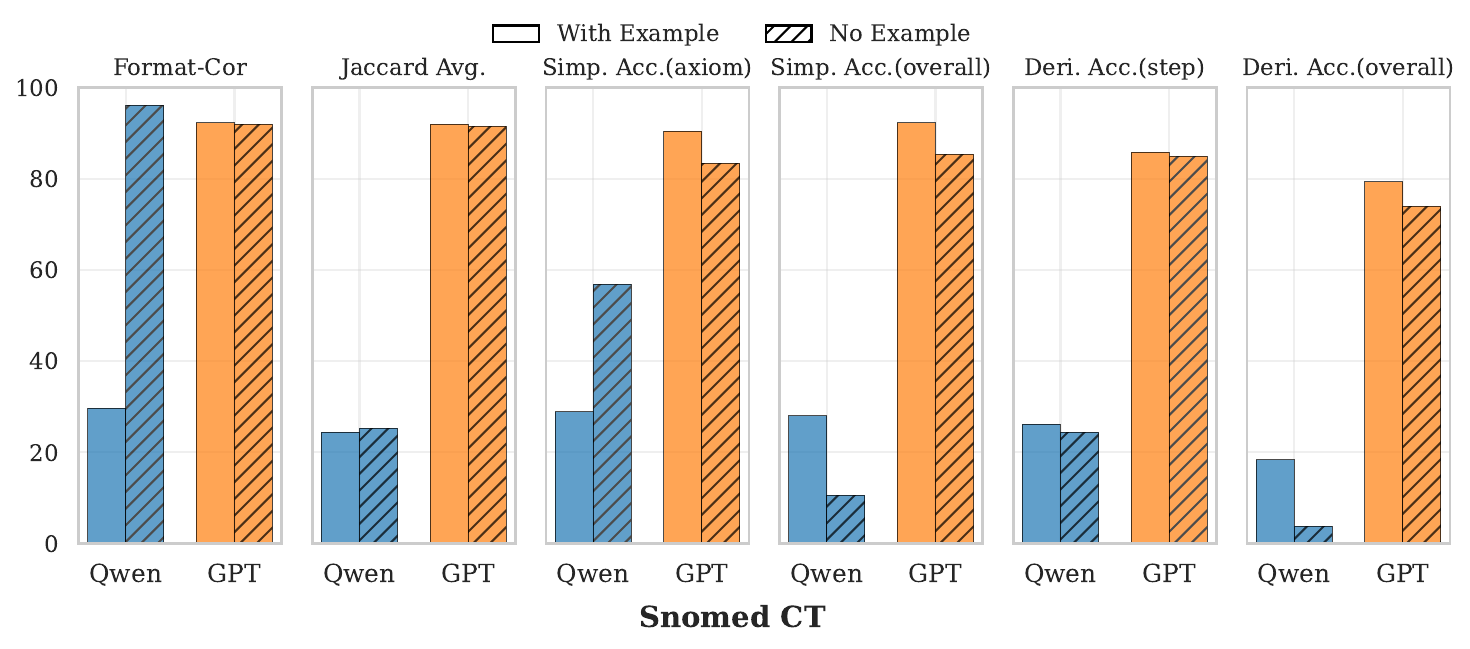}
    \end{subfigure}
    \begin{subfigure}{\linewidth}
        \centering
        \includegraphics[width=0.85\linewidth]{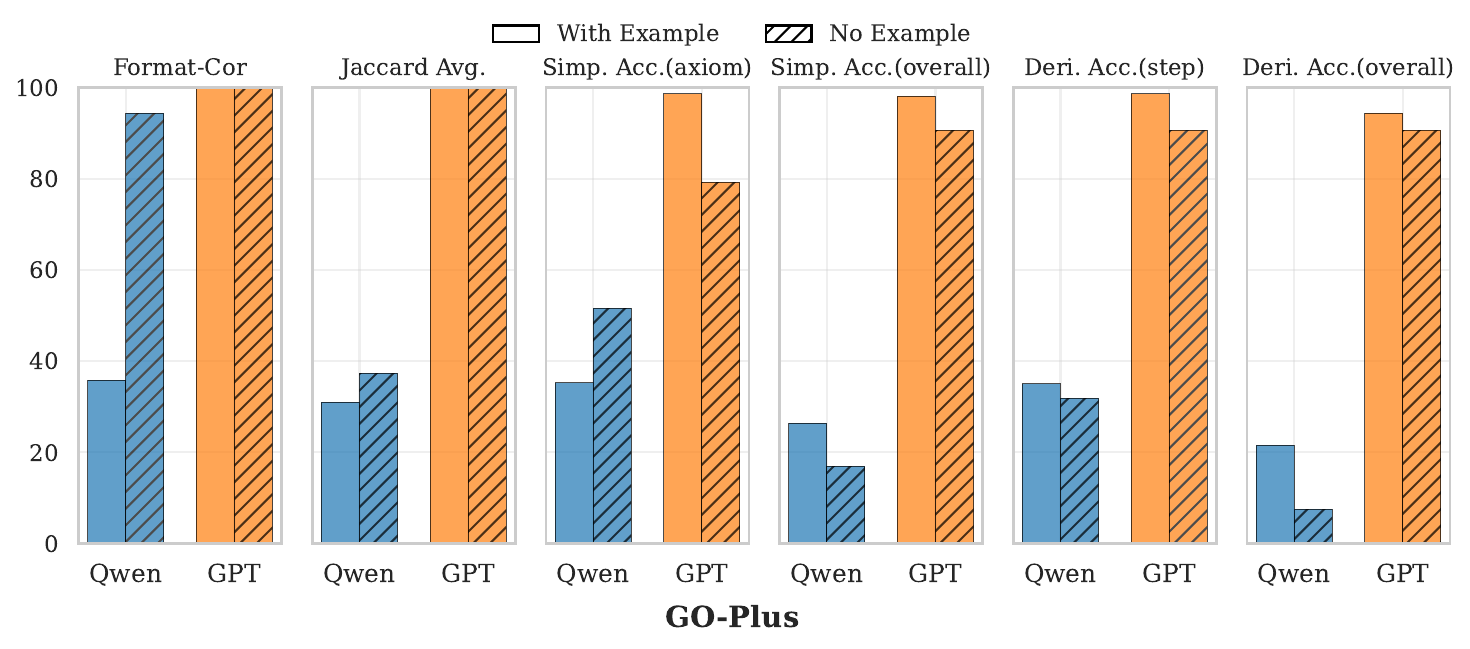}
    \end{subfigure}
        \begin{subfigure}{\linewidth}
        \centering
        \includegraphics[width=0.85\linewidth]{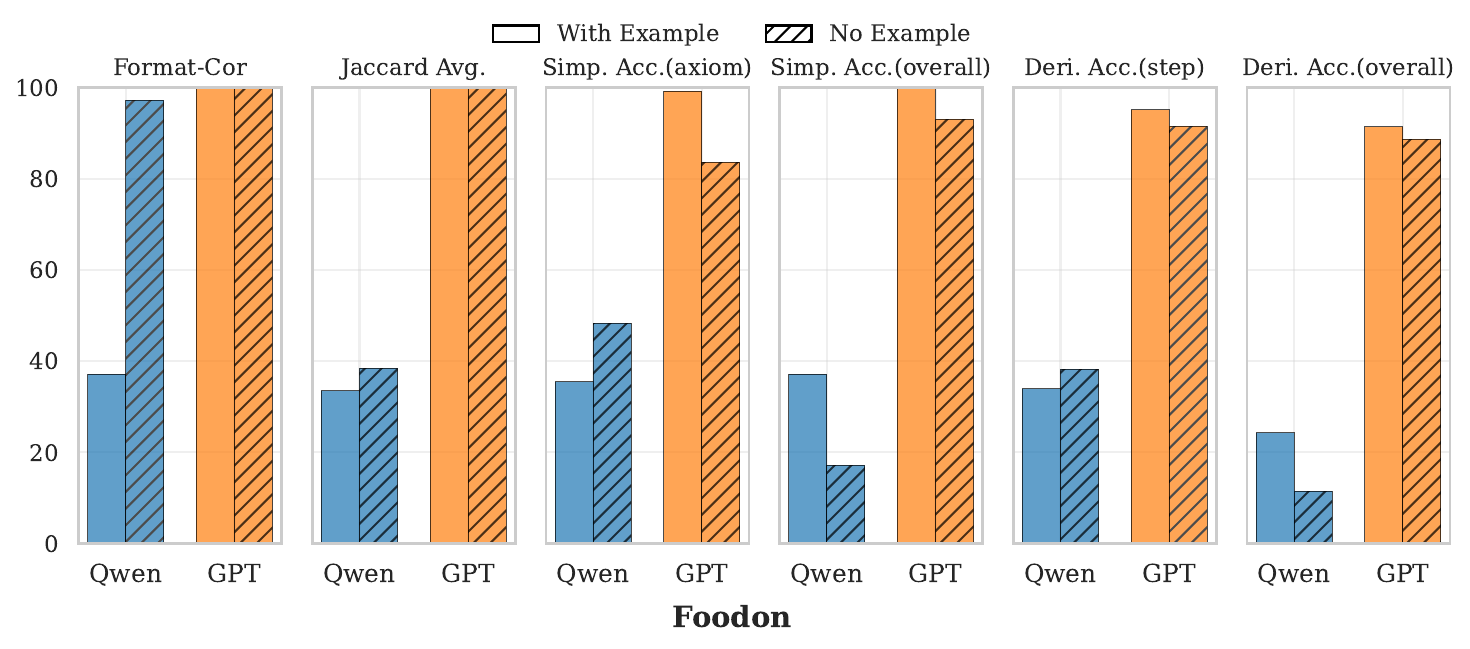}
    \end{subfigure}
    \caption{Comparison (weighted) of \textit{Qwen3-32B} and \textit{GPT-o4-mini} with/without examples.}
    \label{fig:impact_ex}
\end{figure}

\subsubsection{Impact of Inference Rules}\label{sec:impact_ir}  

We now examine the effect of incorporating inference rules, focusing on the two best-performing models: \textit{Qwen3-32B} and \textit{GPT-o4-mini}. The results are presented in Figure~\ref{fig:impact_ir}. 
To ensure a fair comparison that accounts for incorrectly formatted outputs, \textbf{in the rest of the paper, we focus on the weighted values} (recall Remark~\ref{rem:weighted_value}) rather than the raw values reported in Table~\ref{tab:main_result}, unless stated otherwise.

For \textit{Qwen3-32B}, the addition of inference rules yields consistently strong benefits across all ontologies. Both the format correctness rate and the task-specific metrics---Jaccard average, simplification accuracy, and derivation accuracy---improve substantially, with gains of roughly two- to threefold.  

In contrast, the effect of inference rules on \textit{GPT-o4-mini} is much smaller. In the Snomed CT dataset, performance even declines across all metrics of success and accuracy.   
This suggests that \textit{GPT-o4-mini} has largely internalized the relevant inference rules during training. As a result, explicitly encoding these rules in the prompt may instead degrade performance due to the increased input length.


\subsubsection{Impact of Examples}\label{sec:impact_ex}  
We investigate the effect of providing examples, still focusing on \textit{Qwen3-32B}, \textit{GPT-o4-mini}, and their corresponding weighted values. Results are illustrated in Figure~\ref{fig:impact_ex}.

For \textit{Qwen3-32B}, the no-example setting yields a substantially higher rate of format correctness compared to the setting with examples. However, most of the outputs in the no-example case appear to be only formally correct but not logically valid: the case with examples consistently achieves higher overall accuracy across all three tasks, as reflected in higher Jaccard similarity, simplification and derivation overall accuracy. Specifically, for weighted overall derivation accuracy, adding examples to the prompt can increase performance by up to 16-fold (e.g., from 3.82\% to 61.95\% on Snomed CT).  
Interestingly, for axiom-wise simplification accuracy, the case without examples performs consistently better than the case with examples. This may be due to the larger number of correctly formatted outputs in the no-example case, which contribute many correct axiom-wise simplifications and thus increase axiom-wise accuracy.

For \textit{GPT-o4-mini}, the effect of examples is more uniform: performance improves consistently across all metrics. However, the difference is much smaller than in the case of \textit{Qwen3-32B}.

\begin{figure}
    \centering
    \begin{subfigure}[t]{0.445\linewidth}
        \centering
        \includegraphics[width=\linewidth]{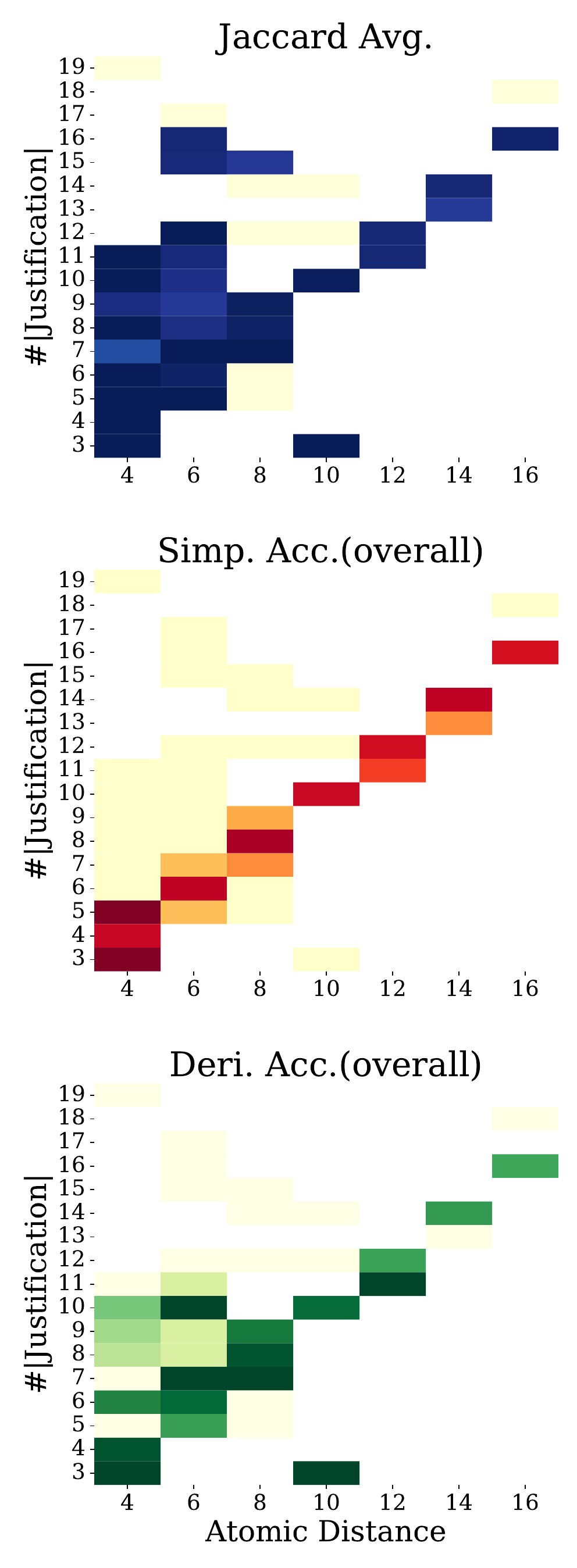}
        \caption{\textit{Qwen3-32B}}
        \label{fig:detailed_illus_overall_snomedct}
    \end{subfigure}
    \begin{subfigure}[t]{0.54\linewidth}
        \centering
        \includegraphics[width=\linewidth]{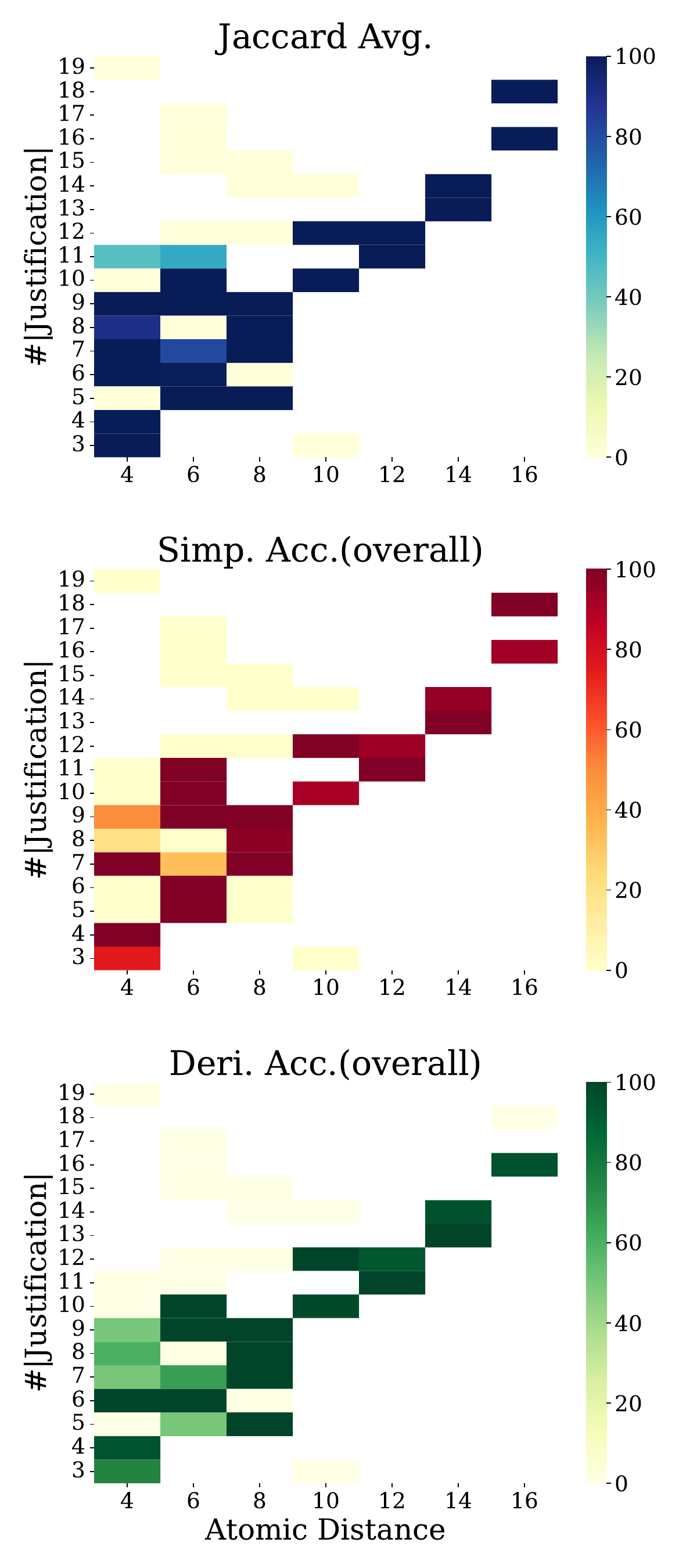}
        \caption{\textit{GPT-o4-mini}}
        \label{fig:detailed_illus_o4-mini-snomedct}
    \end{subfigure}
    \vspace{-0.35cm}
    \caption{Performance (weighted) on Snomed CT across different subsumptions grouped by \subdist and justifications size (with examples and inference rules). }
    \label{fig:detail_snomedct_heatmap}
\end{figure}

\subsubsection{Performance on Different Conclusions} 
Figure~\ref{fig:detail_snomedct_heatmap} provides a fine-grained illustration of the tested conclusions over Snomed CT, grouped by their \subdist\ ($y$-axis) and the number of axioms in the corresponding justifications ($x$-axis).  Results on other datasets are provided in Appendix~\ref{app:more_result}.


Recall from Remark~\ref{rem:subdits} that cases in the upper-left region of Figure~\ref{fig:detail_snomedct_heatmap}—characterized by low \subdist\ but large justification size—are relatively complex and often require intricate reasoning. We observe that the two best-performing LLMs, \textit{GPT-o4-mini} and \textit{Qwen3-32B}, exhibit limited effectiveness on these cases, even when provided with examples and inference rules (additional tests for such difficult cases are reported in Appendix~\ref{app:hard}). 
By contrast, in the diagonal region—where \subdist\ and justification size are roughly equal—both models perform consistently well, regardless of justification size.

\subsection{Complex Settings}\label{sec:complex_case}
To further identify the main challenges faced by LLMs in the proving scenario, we investigate the performance of the best-performing model, \textit{GPT-o4-mini}, under more challenging and realistic conditions. These include varying the \textit{ratio of noisy axioms}, using \textit{natural-language} instead of formal logic representations in the input, and handling \textit{incomplete cases} where the provided axioms are insufficient to derive the given conclusion. We focus on the Snomed CT case, as its corresponding samples are more complex and therefore more representative than those of Foodon and GO-Plus.

\subsubsection{Different Ratios of Noisy Axioms}
We varied the ratio of justification and noisy axioms from 1:1 to 1:5, 1:10, and 1:20 and evaluated the LLMs without examples or inference rules in the prompt. As shown in Table~\ref{tab:o4-mini-complex}, performance consistently declines as the proportion of noise increases, indicate that noisy axioms have a strong negative impact on the model’s final performance. Specifically, as the ratio increases from 1:1 to 1:20, the weighted Jaccard similarity, simplification and derivation overall accuracy fall by around 47\%, 42\%, and 53\%, respectively.

\subsubsection{Natural Language Form}\label{sec:nature_language}

We next investigate performance when axioms are expressed in natural language rather than formal logical notation. For example, instead of the logical statement:
\(
``\texttt{Teacher} \equiv \texttt{Person} \sqcap \exists_{\texttt{teach}}. \texttt{Course} \sqcap \exists_{\texttt{workat}}. \texttt{School}'',
\) 
we use the equivalent natural language description:  
\emph{``Teacher is equivalent to a person who teaches some courses and works at some school.''}  
In this setting, concepts (e.g., \texttt{Teacher}, \texttt{Person}, \texttt{Course}, \texttt{School}) and roles (e.g., \texttt{teach}, \texttt{workat}) are introduced through natural language rather than predefined formal symbols. It is worth noted that in this case it is no longer possible to automatically verify the logical correctness of simplification and derivation steps as all axioms are expressed in natural language. Instead, we focusing on the format correctness rate and Jaccard similarity.

We compare the standard setting (i.e., a true-to-noisy axiom ratio of 1:1) with a variant where axioms are replaced by their natural language forms, as described above. The results are shown on the right-hand side of Table~\ref{tab:o4-mini-complex}, where we also report results with 10,000 max token limit instead of the 5,000 limit by default, since natural-language expressions are substantially longer than formal logic ones. Compared with the 1:1 ratio case in the second column of Table~\ref{tab:o4-mini-complex}, we find that performance with natural-language forms is close to that with formal logic forms (less than an 8\% drop), and increasing the token limit has only a minor effect (less than a 2\% difference). This suggests that the main bottleneck lies in reasoning complexity rather than representation length or expression form.

\begin{table}[tb]
\centering
\caption{Performance (not weighted) under different ratios of justification and noisy axioms (left), and using natural language forms with a justification-noise ratio of 1:1 (right). Results are from Snomed CT using \textit{GPT-o4-mini} without inference rules or examples.}
\vspace{-0.35cm}
\label{tab:o4-mini-complex}
\resizebox{0.48\textwidth}{!}{
\begin{tabular}{p{0.25cm}lcccc|cccc}
\toprule
\multirow{3}{*}{\textbf{Metric}} & &\multicolumn{4}{c|}{\textbf{Ratio}} & \multicolumn{2}{|c}{\textbf{Nature Lang.}}\\
 & & \multicolumn{4}{c|}{(justification : noise)} &\multicolumn{2}{|c}{(max token)} \\
\cmidrule(lr){3-6} \cmidrule(lr){7-8}
&& \textbf{1 : 1} & \textbf{1 : 5} & \textbf{1 : 10}& \textbf{1 : 20} & 5000 & 10,000 \\
\midrule
\multicolumn{2}{l}{\textbf{Format-Correct}} & 91.98& 80.57& 78.28 & 61.71 & 84.00& 85.42\\
\midrule
\multicolumn{2}{l}{\textbf{Jaccard Avg.}} & 99.49 & 95.49 &87.71& 78.96 & 96.00 &94.56 \\
\midrule
\multirow{2}{=}{\raggedright \textbf{Simp. Acc.}} 
    & \hspace{1em}axiom-wise & 90.60 & 92.86 & 91.76 &89.74 & -& -\\
    & \hspace{1em}overall & 80.91 & 77.99 & 72.27 &69.29 & -& - \\
\midrule
\multirow{2}{=}{\raggedright \textbf{Deri. Acc.}} 
    & \hspace{1em}step-wise & 92.33  & 93.87 & 90.50 &88.60& -& -\\
    & \hspace{1em}overall & 80.30 & 73.13 & 64.30 &56.00& - & -\\
\bottomrule
\end{tabular}}
\end{table}

\begin{figure}
    \centering
    \includegraphics[width=\linewidth]{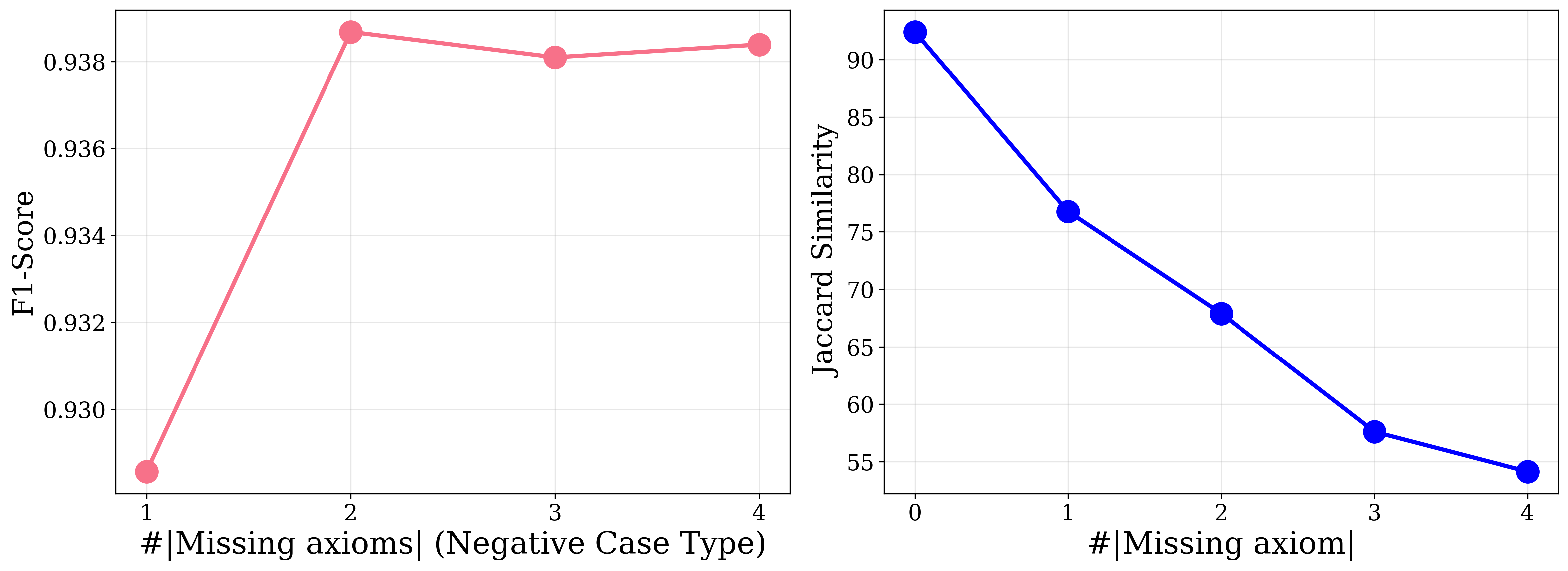}
   \caption{F1-scores for recognizing logical completeness (left) and weighted Jaccard similarity for identifying axioms belonging to justifications (right). Results are from \textit{GPT-o4-mini} without inference rules or examples.}
    \label{fig:missing-case}
\end{figure}

\subsubsection{Incomplete Premises}

We evaluate LLM performance in cases where the given axioms are insufficient to derive the conclusion. To fit such incomplete settings, we use natural language formulations instead of symbolic representations, enabling richer semantic interpretation. We consider two tasks:   
(1) assessing whether the axioms are logically complete with respect to the conclusion (\emph{Logic Completeness}); and  
(2) identifying axioms that appear in at least one valid justification of the conclusion.

Incomplete samples are created by removing subsets of axioms from their justifications in Snomed CT examples whose conclusions have an \subdist greater than 9.   
For task (1), the original samples serve as positive instances, while a 1:1 set of negative instances is generated by removing $k$ ($1\leq k\leq 4$) axioms from each. Performance is measured by the F1-score.  
For task (2), we use the negative samples above, regard the remaining justification axioms as ground truth, and evaluate LLMs using the Jaccard Similarity metric.

The results, summarized in Figure~\ref{fig:missing-case}, reveal two main trends. First, the high F1-scores computed across sets with different numbers of missing axioms in the negative samples (all around $0.93$ when $k$ ranges from $1$ to $4$)  
indicate that the models are generally effective at detecting incomplete premises.
Second, the ability to identify the remaining relevant axioms declines sharply up to 38\% as more axioms are removed, as evidenced by the decrease in mean Jaccard similarity. This suggests that higher degrees of incompleteness significantly impair the model’s capacity to identify relevant information, even when semantic cues are available in the input.






\section{Conclusion and Future Work}
We evaluated the use of LLMs for generating proofs in OWL ontologies, including extraction, simplification, and explanation tasks, as well as handling challenging cases involving incomplete premises or natural language formulations. We developed a method for automatic construction of logic proving datasets with different settings from OWL ontologies, and evaluated $7$ recent reasoning LLMs. Our findings indicate that while LLMs achieve promising results overall, they still struggle in complex scenarios, with substantial performance degradation facing complex logic patterns, noisy axioms or incomplete premises.

For future work, we aim to extend our evaluation to more complex fragments or the full OWL 2 DL using broader sampling strategies, and to more complex reasoning tasks, such as the computation of deductive modules~\cite{DBLP:conf/ijcai/YangK0B23, grau2008modular} or reasoning beyond subsumption.
With these evaluation results, we are also interested in developing LLM-based frameworks to retrieve explainable knowledge from OWL ontologies for augmenting generation.

\bibliographystyle{ACM-Reference-Format}
\bibliography{sample-base}

\appendix
\newpage

\section{Implementation Details}\label{app:implementation}

For the open-source LLMs, we conducted experiments locally on a machine equipped with a single A100 80 GB GPU, leveraging the \texttt{vllm}~\cite{kwon2023efficient} package to accelerate inference. For \textit{GPT-o4-mini}, we accessed the model through the ChatGPT API, with an average batch-mode cost of under \$0.01 per request. 
Our implementation is based on the DeepOnto package \cite{he2024deeponto}. We use the ELK reasoner \cite{kazakov2012elk} to compute justifications and to verify whether each derivation step is correct.

\subsection{Standard Case}
The prompt used for evaluating the performance of LLMs on three tasks involving explanation and proof generation based on OWL ontologies is shown in Figure \ref{fig:prompt-full}. 
The input consists of all axioms (each indexed) along with the desired conclusion. The output is divided into three sections: \texttt{AXIOMS\_USED}, \texttt{SIMPLIFY}, and \texttt{DERIVE}:  
\begin{enumerate}[leftmargin=*] 
    \item \texttt{AXIOMS\_USED} specifies the indices of the axioms required to derive the given conclusion. In the given example, all axioms except for the last one are needed.  
    
    \item \texttt{SIMPLIFY} presents the simplified forms of the selected axioms that are sufficient for deriving the conclusion. For instance, in the given example, instead of using the full first axiom $A \equiv \exists r. B$, it is sufficient to use the weaker version $\exists r. B \sqsubseteq A$ to derive the given conclusion, which is a direct consequence of the original form.
    
    \item \texttt{DERIVE} provides a step-by-step explanation of the reasoning process. The derivation is split into distinct stages, with each stage ending in an crucial intermediate result  for inferring the final conclusion. For example, in the given case, $D \sqsubseteq A$ is chosen as the key intermediate result, and the derivation is correspondingly divided into two steps.
\end{enumerate}

\begin{figure}[h]
    \centering
  
    \footnotesize
    \begin{fullpromptbox}

\texttt{\# Explanation of Logic Inference} \\
\texttt{You are an expert in logical reasoning. Your task is to produce a clear and easily understandable explanation that demonstrates how the conclusion logically follows from the given axioms.}

\vspace{0.8em}
{\color{red}\texttt{(Inference rules here if applied)}}

\vspace{0.8em}
\texttt{\#\# Output Format Requirements} \\
\texttt{Your response MUST follow this exact format:}

\vspace{0.5em}
\texttt{AXIOMS\_USED: \{list of axiom identifiers\}}

\vspace{0.5em}
\texttt{SIMPLIFY:}\\
\texttt{[axiom\_id]}: \texttt{[original\_axiom]} $\rightarrow$ \texttt{[simplified\_form]}\\
\texttt{...}\\

\vspace{0.5em}
\texttt{DERIVE:}\\
\texttt{STEP[n]}: \texttt{[premise\_list]} $\vdash$ \texttt{[conclusion]}\\
\texttt{EXPLANATION:} \texttt{[a simple explanation of the derivation]}\\
\texttt{...}

\vspace{0.8em}
{\color{red}\texttt{(An example here if applied)}}

\vspace{0.8em}
\texttt{\#\# Task \\
\texttt{Given the following axioms:}
\begin{itemize}[leftmargin=2em]
  \item[] (0) {\color{blue} Axiom 0}
  \item[] (1)  {\color{blue} Axiom 1}
  \item[] $\ldots$
\end{itemize}}

\vspace{0.5em}
\texttt{**Target Conclusion**: \{{\color{blue}conclusion}\}}

\vspace{0.8em}
\texttt{**Instructions:**
\begin{enumerate}[leftmargin=2em]
  \item Identify and list only the axioms that are essential for deriving the conclusion. Note that only half of the given axioms are necessary for the derivation.
  \item Simplify each selected axiom by extracting only the logically relevant components.
  \item Present your response using the required DSL format, with clearly defined \texttt{AXIOMS\_USED}, \texttt{SIMPLIFY}, and \texttt{DERIVE} sections.
  \item In the \texttt{DERIVE} section, break the reasoning into meaningful steps. Each \texttt{STEP} should capture a key intermediate result that contributes to the overall derivation, while maintaining logical clarity and completeness.
\end{enumerate}
}

\end{fullpromptbox}

\caption{Prompt for logical explanation and proof (input shown in blue; placeholders for inference rules or examples shown in red).}
    \label{fig:prompt-full}
\end{figure}

Note that we output all three tasks simultaneously rather than handling them one by one, since their underlying requirement is the same: each demands a detailed analysis of the reasoning process needed to derive the given conclusion. However, the difficulty increases progressively across the tasks. First, we ask only for the indices of the axioms required. Next, we require the LLM to identify the relevant portions of those axioms that are actually useful. Finally, we ask for a complete derivation process, structured as an explanation segmented by crucial intermediate results.

\subsection{Incomplete case}
In the incomplete case, where the input axioms may not sufficient to infer the given conclusion, we focus on the natural language forms and therefore omit the \texttt{SIMPLIFY} and \texttt{DERIVE} parts from the standard case, as well as the inference rules and examples. The adjusted prompt is presented in Figure \ref{fig:prompt-missing}, with the differences in the required output and instructions highlighted in red.

\begin{figure}
    \centering
    \footnotesize
    \begin{fullpromptbox}

\texttt{\# Logical Inference Task} \\
\texttt{You are an expert in logical reasoning. Your task is to assess whether the target conclusion logically follows from the provided axioms, and to identify any potentially missing axioms if it does not.}

\vspace{0.8em}

\texttt{\#\# Required Output Format} \\
\texttt{Your response MUST strictly follow this format:}

\vspace{0.5em}

\textcolor{orange}{
\texttt{MISSING: [YES/NO]} \\
\texttt{AXIOMS\_USEFUL: [list of relevant axiom identifiers]} \\
\texttt{SUSPECTED\_MISSING\_PARTS: NONE / [potential missing knowledge in natural language]}
}

\vspace{0.8em}

\texttt{\#\# Input Data} \\
\texttt{Consider the following axioms:}
\begin{itemize}[leftmargin=2em]
  \item[(0)] {\color{blue} Axiom 0}
  \item[(1)] {\color{blue} Axiom 1}
  \item[$\ldots$]
\end{itemize}

\vspace{0.5em}

\texttt{**Target Conclusion**: \{{\color{blue} conclusion}\}}

\vspace{0.8em}

\texttt{\#\# Instructions} \\
\textcolor{orange}{
\texttt{1. Decide whether the axioms are sufficient to derive the target conclusion.} \\
\texttt{2. In AXIOMS\_USEFUL, include only the axioms that contribute to the derivation of the target conclusion.} \\
\texttt{3. If MISSING = YES, provide the suspected missing axioms in SUSPECTED\_MISSING\_PARTS.} \\
\texttt{4. End your response with exactly the three required fields (MISSING, AXIOMS\_USEFUL, SUSPECTED\_MISSING\_PARTS) and nothing else.}
}

\end{fullpromptbox}
\caption{Prompt for the incomplete case, with inputs shown in blue and differences from the standard case highlighted in orange.}
    \label{fig:prompt-missing}
\end{figure}

\begin{table}
\centering
\caption{Unweighted performance of GPT-o4-mini (without inference rules or examples) on the SNOMED CT dataset. “Original” refers to the SNOMED CT data used in Table~\ref{tab:main_result}, differences computed by weighted values are indicated by $^*$.}
\label{tab:o4-mini-transposed}
\begin{tabular}{lcc|c}
\toprule
\textbf{Metric} & \textbf{Original} & \textbf{Hard} & \textbf{Difference} \\
\midrule
\textbf{Format-Correct} & 91.98 & 51.72 & $-43.77\%$ \\
\midrule
\textbf{Jaccard Avg.} & 99.49 & 94.39 & $-42.70\%^*$ \\
\midrule
\textbf{Simp. Acc. (axiom-wise)} & 90.60 & 87.62 & $-38.01\%^*$  \\
\textbf{Simp. Acc. (overall)} & 80.91 & 40.14 & $-53.66\%^*$  \\
\midrule
\textbf{Deri. Acc. (step-wise)} & 92.33 & 79.69 & $-43.70\%^*$  \\
\textbf{Deri. Acc. (overall)} & 80.30 & 32.39 & $-57.11\%^*$  \\
\bottomrule
\end{tabular}
\end{table}

\section{Other results}\label{app:other_results}

\subsection{Hard case with GPT-o4-mini}\label{app:hard}
\begin{figure}
    \centering
   \includegraphics[width=\linewidth]{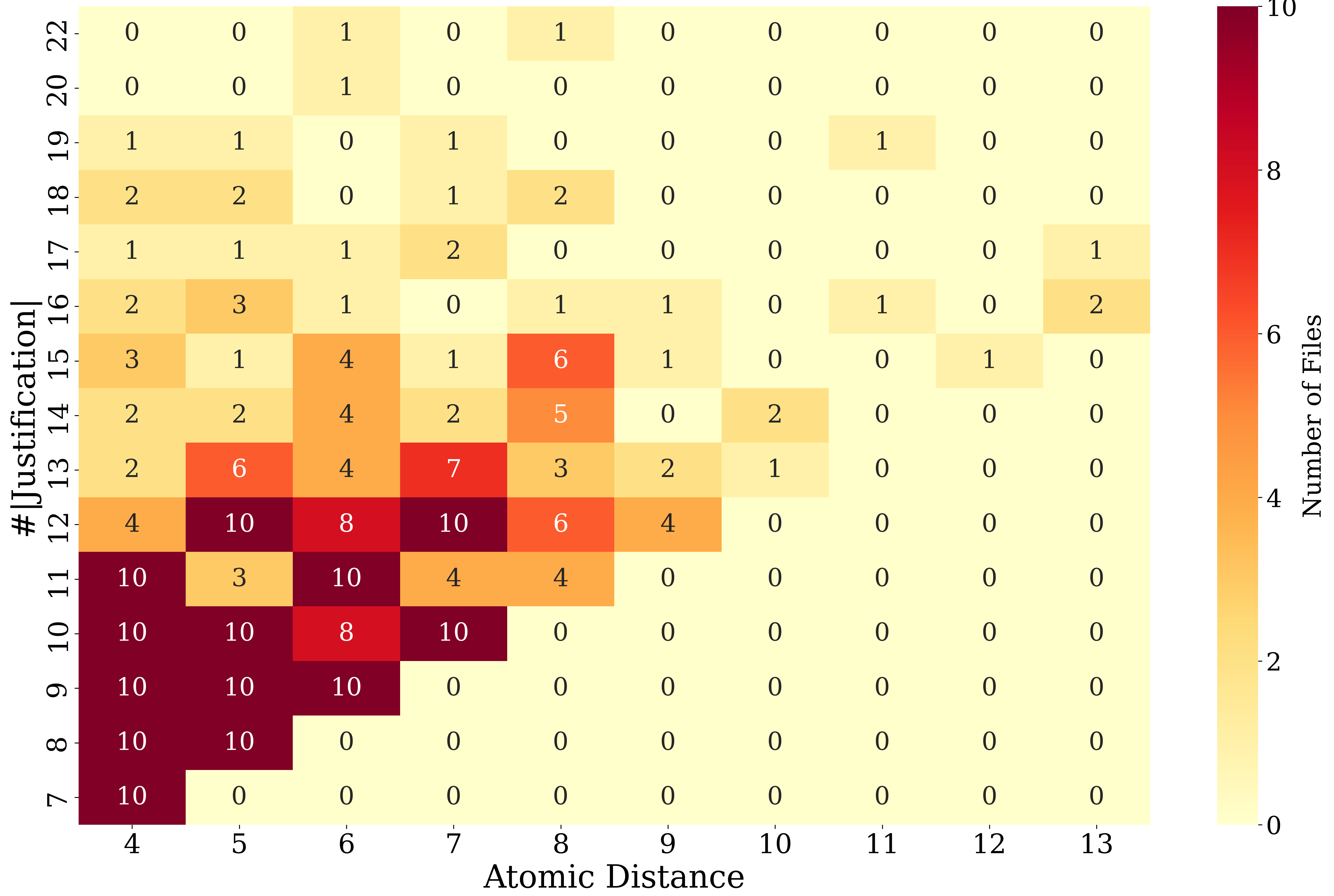}
      \caption{Statistics of the hard cases extracted from Snomed CT.}
    \label{fig:hard_statistic}
\end{figure}

The number of hard cases—i.e., subsumptions with large \subdist\ but small justification size—in the original dataset is relatively small. To enable more reliable analysis, we further analyzed the hard cases, comprising 261 additional subsumptions in Snomed CT, where the justification size exceeds their \subdist by at least 3. Specifically, we considered 300 subsumptions for each subsumption group with fixed \subdist k, where k ranging from 4 to 20, and then selected only those whose justification size exceeds \subdist at least 3. To further control the distribution, we limited the sampling to at most 10 subsumptions for each (\subdist, \#|Justification|) group. The detailed statistics of these hard cases are presented in Figure~\ref{fig:hard_statistic}.

\begin{figure*}[t]
    \centering
        \centering
        \begin{subfigure}[t]{0.23\linewidth}
            \centering
            \includegraphics[width=\linewidth]{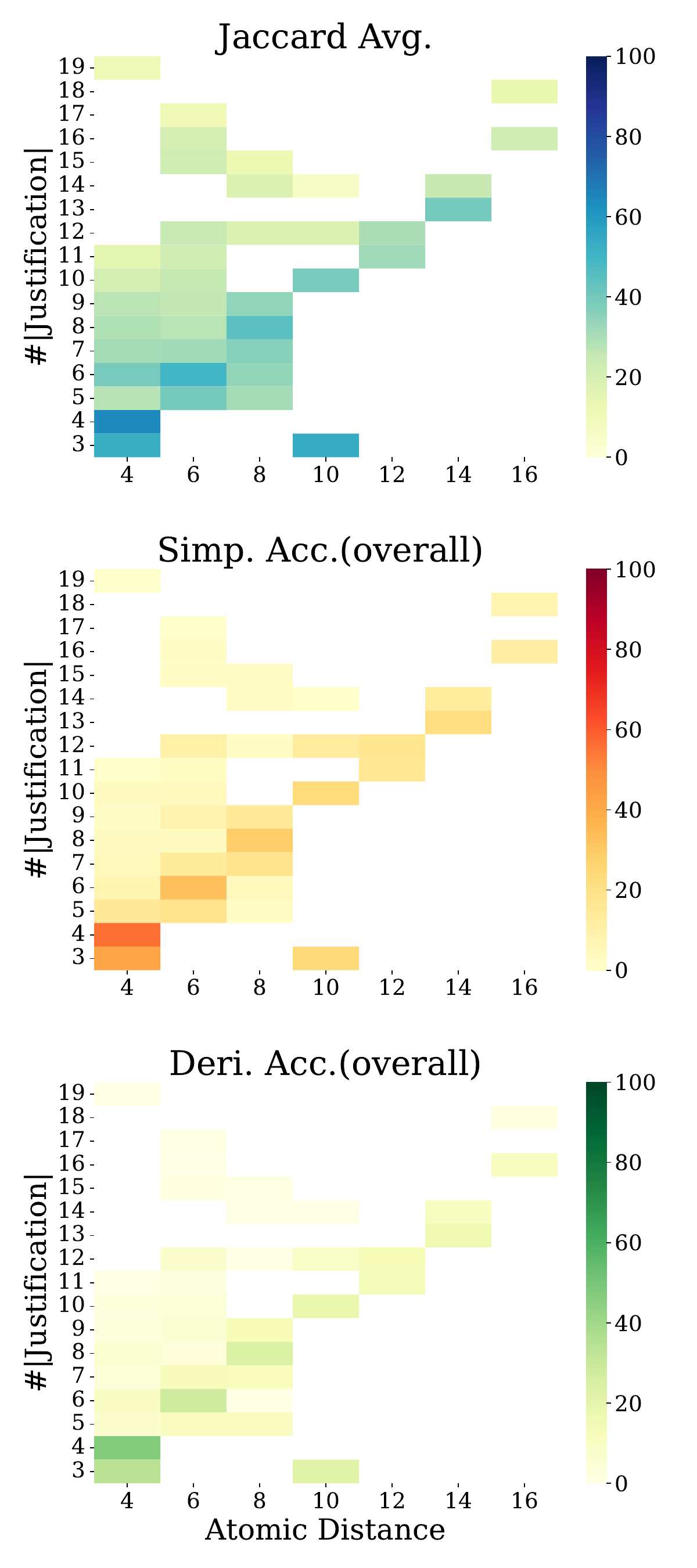}
            \caption{Snomed CT (avg.)}
            \label{fig:overall_snomedct}
        \end{subfigure}%
        \begin{subfigure}[t]{0.23\linewidth}
            \centering
            \includegraphics[width=\linewidth]{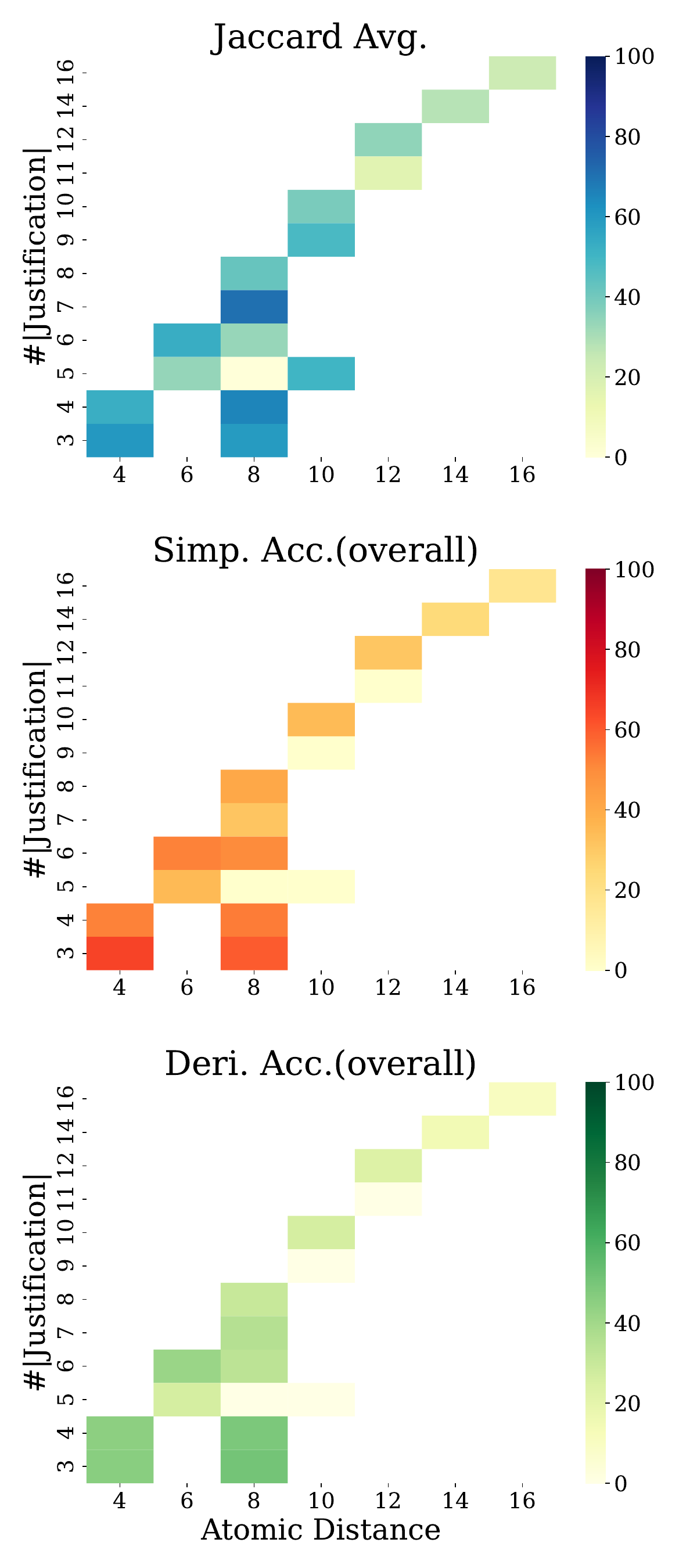}
            \caption{Foodon (avg.)}
            \label{fig:overall_Foodon}
        \end{subfigure}%
        \begin{subfigure}[t]{0.23\linewidth}
            \centering
            \includegraphics[width=\linewidth]{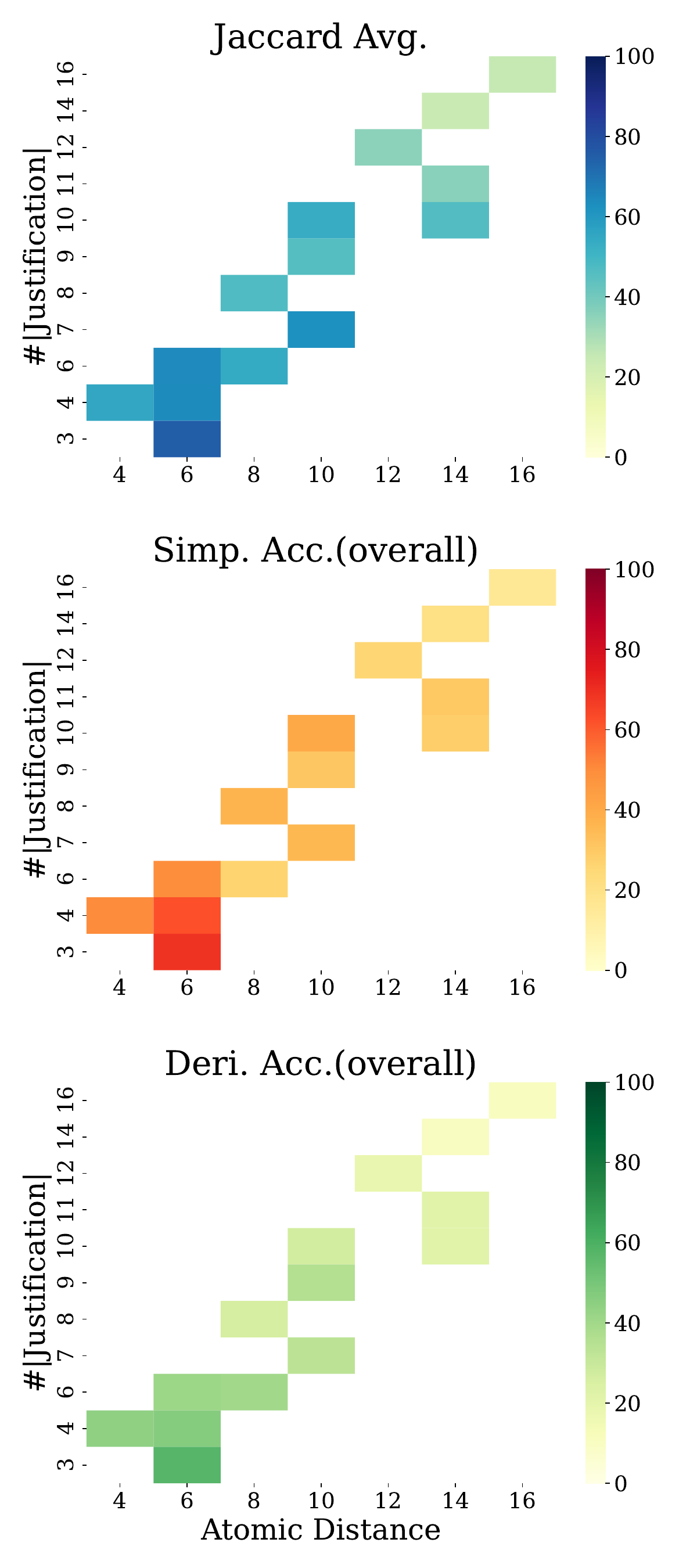}
            \caption{GO-Plus (avg.)}
            \label{fig:overall_goplus}
        \end{subfigure}
          \begin{subfigure}[t]{0.23\linewidth}
            \centering
            \includegraphics[width=\linewidth]{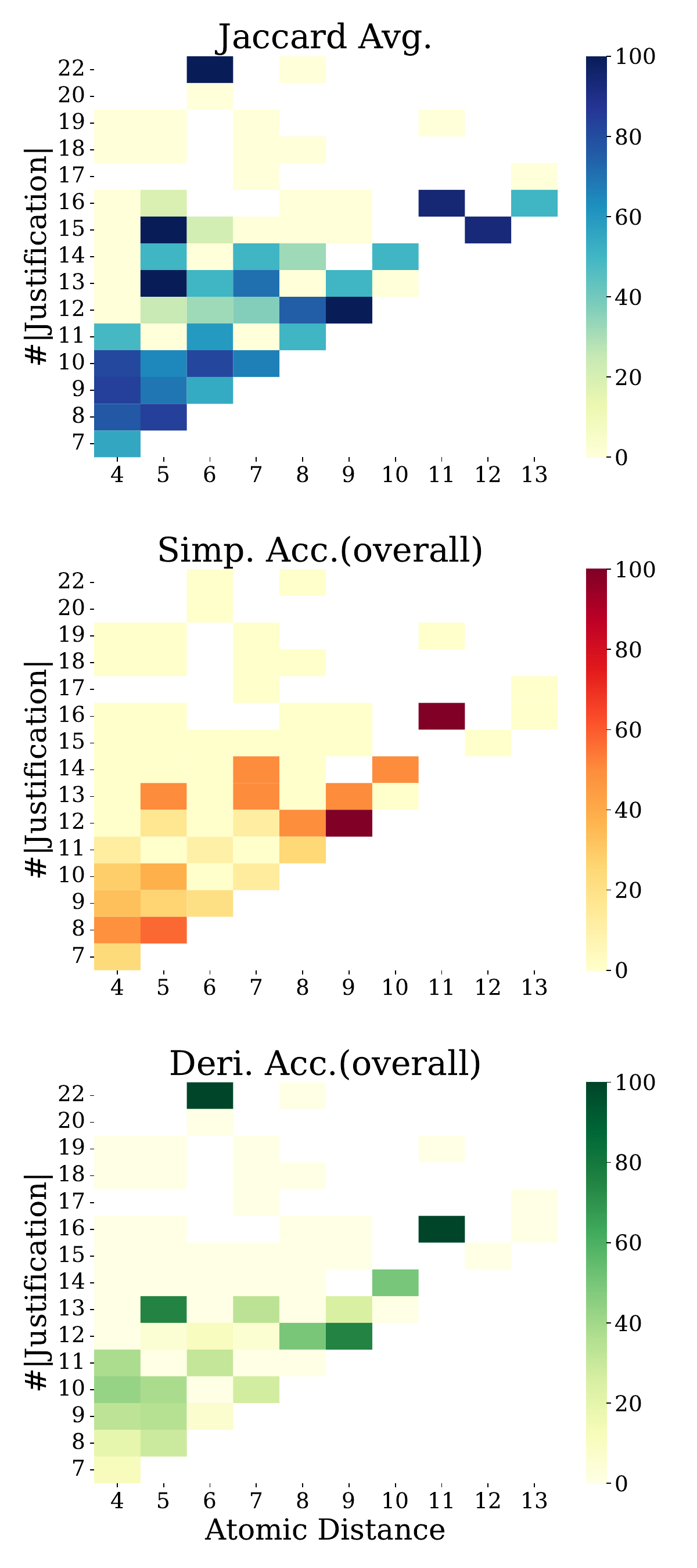}
        \caption{Hard (\textit{GPT-o4-mini})}
        \label{fig:hard_performance}
        \end{subfigure}%
        \caption{Average performance (weighted) of all methods on \textbf{Foodon} and \textbf{GO-Plus}, and the performance of \textit{GPT-o4-mini} (without example or inference rules) on the Hard dataset, measured across subsumption distance and justification size.}
        \label{fig:combined_overall_Foodon_goplus}
\end{figure*}

We then evaluated GPT-o4-mini under the base setting (i.e., without inference rules or in-context examples) on this dataset. The overall results are summarized in Table~\ref{tab:o4-mini-transposed}.
We observed a substantial drop in both the format correctness rate and overall accuracy by nearly half, which prove the limitation of LLMs on those hard cases.  The detailed performance across different (\subdist, \#|Justification|) groups is shown in Figure~\ref{fig:hard_performance}.

\subsection{Statistics of Subsumptions}
\begin{figure}
        \centering
    \includegraphics[width=0.9\linewidth]{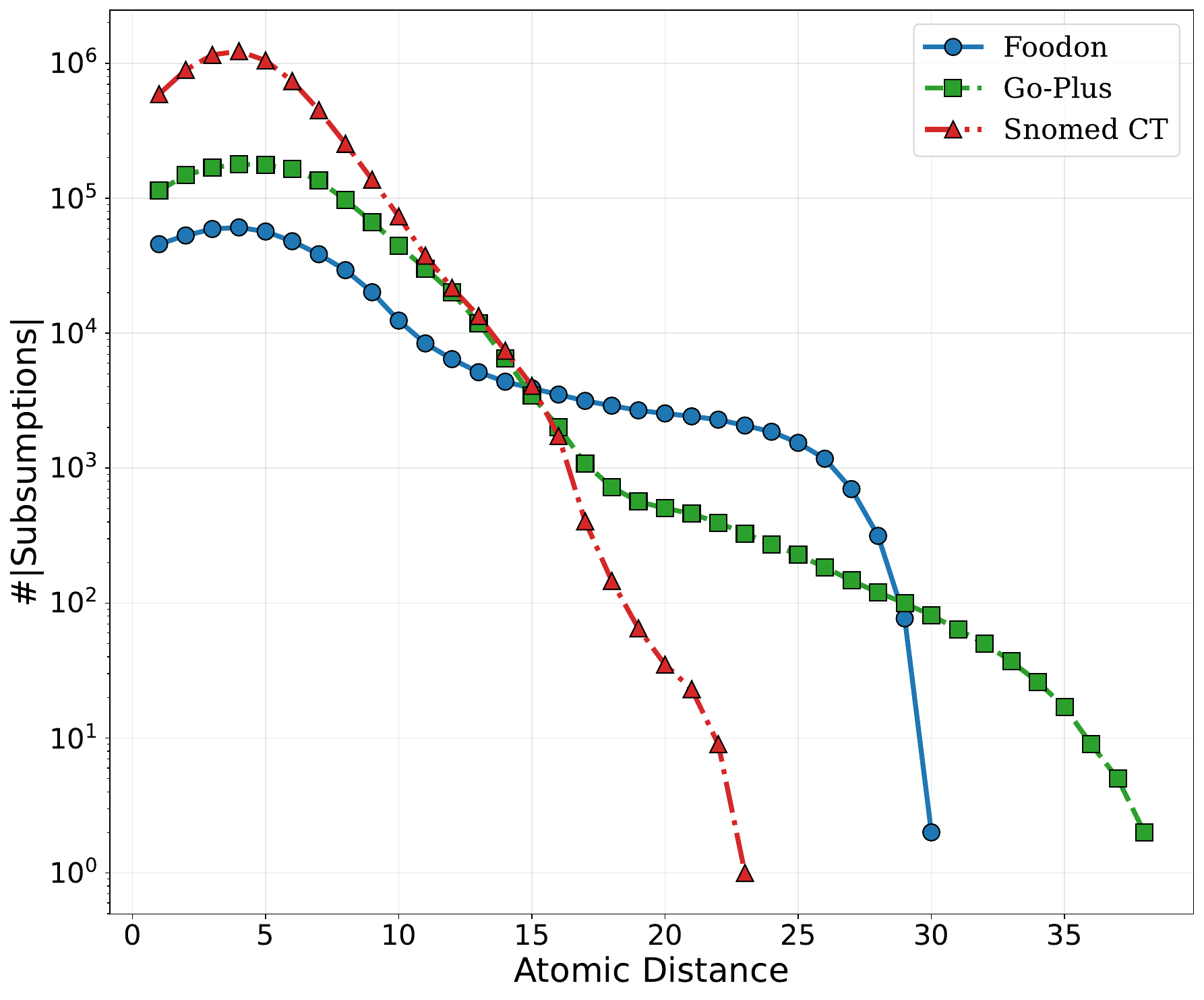}
    \caption{ Number of all subsumptions with different  \subdist.}
    \label{fig:stat_subdist}
\end{figure}

Here, in Figure~\ref{fig:stat_subdist}, we present detailed statistics of the subsumptions derived from the Snomed CT, GO-Plus, and Foodon ontologies. The distribution of \subdist\ ranges from 0 to 38, with up to millions of distinct subsumptions sharing the same \subdist\ value. Moreover, for \subdist\ values below 15, there are consistently at least several thousand corresponding subsumptions. This provides a rich resource for further evaluations, as well as potential investigations or implementations beyond the few hundred samples examined in our study.

\subsection{Average Results Across LLMs} \label{app:more_result}
Figure~\ref{fig:combined_overall_Foodon_goplus} presents the overall results for the SNOMED CT, FoodOn, and GO-Plus ontologies. The FoodOn and GO-Plus datasets appear to be relatively easier in terms of logical explanation complexity, as most instances lie near the diagonal where \subdist\ equals the justification size. This is primarily because most $\mathcal{EL}$ axioms in these ontologies are simple subsumption axioms of the form $A \sqsubseteq B$, with $A$ and $B$ being atomic concepts. Consequently, the average performance of all LLMs across different settings is higher for FoodOn and GO-Plus than for SNOMED CT, further supporting our claim that the SNOMED CT data are comparatively more challenging.

\subsection{Effect of Example and Input in Prompt}

\begin{figure}
    \centering
    \begin{tcolorbox}[title=Example,colback=lightgray,colframe=black]
\texttt{\#\# Example}

\texttt{Given the following axioms:} \\
\texttt{[1] A }$\equiv$ $\exists \texttt{r.B}$ \\
\texttt{[2] C }$\sqsubseteq$\texttt{ B }$\sqcap$\texttt{ H }$\sqcap$\texttt{ I} \\
\texttt{[3] D }$\equiv$$\exists \texttt{r.C}$ $\sqcap$\texttt{ G} \\
\texttt{[4] E }$\equiv$\texttt{ A }$\sqcap$\texttt{ F} \\
\texttt{[5] D }$\sqsubseteq$\texttt{ F }$\sqcap$\texttt{ J}

\vspace{1em}
\texttt{The desired explanation for deriving the conclusion D }$\sqsubseteq$\texttt{ E is as follows:}

\vspace{1em}
\texttt{AXIOMS\_USED: 1,2,3,4,5}

\texttt{SIMPLIFY:} \\
\texttt{[1] A }$\equiv$ $\exists \texttt{r.B}$ \texttt{ → }$\exists \texttt{r.B}$ $\sqsubseteq$\texttt{ A} \\
\texttt{[2] C }$\sqsubseteq$\texttt{ B }$\sqcap$\texttt{ H }$\sqcap$\texttt{ I } \texttt{→ C }$\sqsubseteq$\texttt{ B} \\
\texttt{[3] D }$\equiv$ $\exists \texttt{r.C}$ $\sqcap$\texttt{ G \texttt{→ D }$\sqsubseteq$ $\exists \texttt{r.C}$} \\
\texttt{[4] E }$\equiv$\texttt{ A }$\sqcap$\texttt{ F \texttt{→ A }$\sqcap$\texttt{ F }$\sqsubseteq$\texttt{ E}} \\
\texttt{[5] D }$\sqsubseteq$\texttt{ F }$\sqcap$\texttt{ J \texttt{→ D }$\sqsubseteq$\texttt{ F}}

\vspace{1em}
\texttt{DERIVE:} \\

\texttt{STEP1: [1] }$\vdash$\texttt{ $\exists \texttt{r.B} \sqsubseteq \texttt{A}$} \\
\texttt{    EXPLANATION: From equivalence }$\texttt{A} \equiv \exists \texttt{r.B}$\texttt{, we get }$\exists \texttt{r.B} \sqsubseteq \texttt{A}$

\vspace{0.5em}

\texttt{STEP2: [2] }$\vdash$\texttt{ C $\sqsubseteq$ B} \\
\texttt{    EXPLANATION: From conjunction }$\texttt{C} \sqsubseteq \texttt{B} \sqcap \texttt{H} \sqcap \texttt{I}$\texttt{, we get }$\texttt{C} \sqsubseteq \texttt{B}$

\vspace{0.5em}

\texttt{STEP3: [3] }$\vdash$\texttt{ D $\sqsubseteq \exists \texttt{r.C}$} \\
\texttt{    EXPLANATION: From equivalence }$\texttt{D} \equiv \exists \texttt{r.C} \sqcap \texttt{G}$\texttt{, we get }$\texttt{D} \sqsubseteq \exists \texttt{r.C}$

\vspace{0.5em}

\texttt{STEP4: [STEP2, STEP3] }$\vdash$\texttt{ D $\sqsubseteq \exists \texttt{r.B}$} \\
\texttt{    EXPLANATION: From }$\texttt{D} \sqsubseteq \exists \texttt{r.C}$\texttt{ and }$\texttt{C} \sqsubseteq \texttt{B}$\texttt{, we get }$\texttt{D} \sqsubseteq \exists \texttt{r.B}$

\vspace{0.5em}

\texttt{STEP5: [STEP1, STEP4] }$\vdash$\texttt{ D $\sqsubseteq$ A} \\
\texttt{    EXPLANATION: From }$\exists \texttt{r.B} \sqsubseteq \texttt{A}$\texttt{ and }$\texttt{D} \sqsubseteq \exists \texttt{r.B}$\texttt{, we get }$\texttt{D} \sqsubseteq \texttt{A}$

\vspace{0.5em}

\texttt{STEP6: [5] }$\vdash$\texttt{ D $\sqsubseteq$ F} \\
\texttt{    EXPLANATION: From conjunction }$\texttt{D} \sqsubseteq \texttt{F} \sqcap \texttt{J}$\texttt{, we get }$\texttt{D} \sqsubseteq \texttt{F}$

\vspace{0.5em}

\texttt{STEP7: [STEP5, STEP6] }$\vdash$\texttt{ D $\sqsubseteq$ A $\sqcap$ F} \\
\texttt{    EXPLANATION: From }$\texttt{D} \sqsubseteq \texttt{A}$\texttt{ and }$\texttt{D} \sqsubseteq \texttt{F}$\texttt{, we get }$\texttt{D} \sqsubseteq \texttt{A} \sqcap \texttt{F}$

\vspace{0.5em}

\texttt{STEP8: [4] }$\vdash$\texttt{ A $\sqcap$ F $\sqsubseteq$ E} \\
\texttt{    EXPLANATION: From equivalence }$\texttt{E} \equiv \texttt{A} \sqcap \texttt{F}$\texttt{, we get }$\texttt{A} \sqcap \texttt{F} \sqsubseteq \texttt{E}$

\vspace{0.5em}

\texttt{STEP9: [STEP7, STEP8] }$\vdash$\texttt{ D $\sqsubseteq$ E} \\
\texttt{    EXPLANATION: From }$\texttt{D} \sqsubseteq \texttt{A} \sqcap \texttt{F}$\texttt{ and }$\texttt{A} \sqcap \texttt{F} \sqsubseteq \texttt{E}$\texttt{, we get }$\texttt{D} \sqsubseteq \texttt{E}$

\end{tcolorbox}
    \caption{Example with detailed derivation steps.}
    \label{fig:prompt-example-detail}
\end{figure}




\begin{figure*}[t]
    \centering
    \begin{subfigure}[t]{0.32\linewidth}
        \centering
        \includegraphics[width=\linewidth]{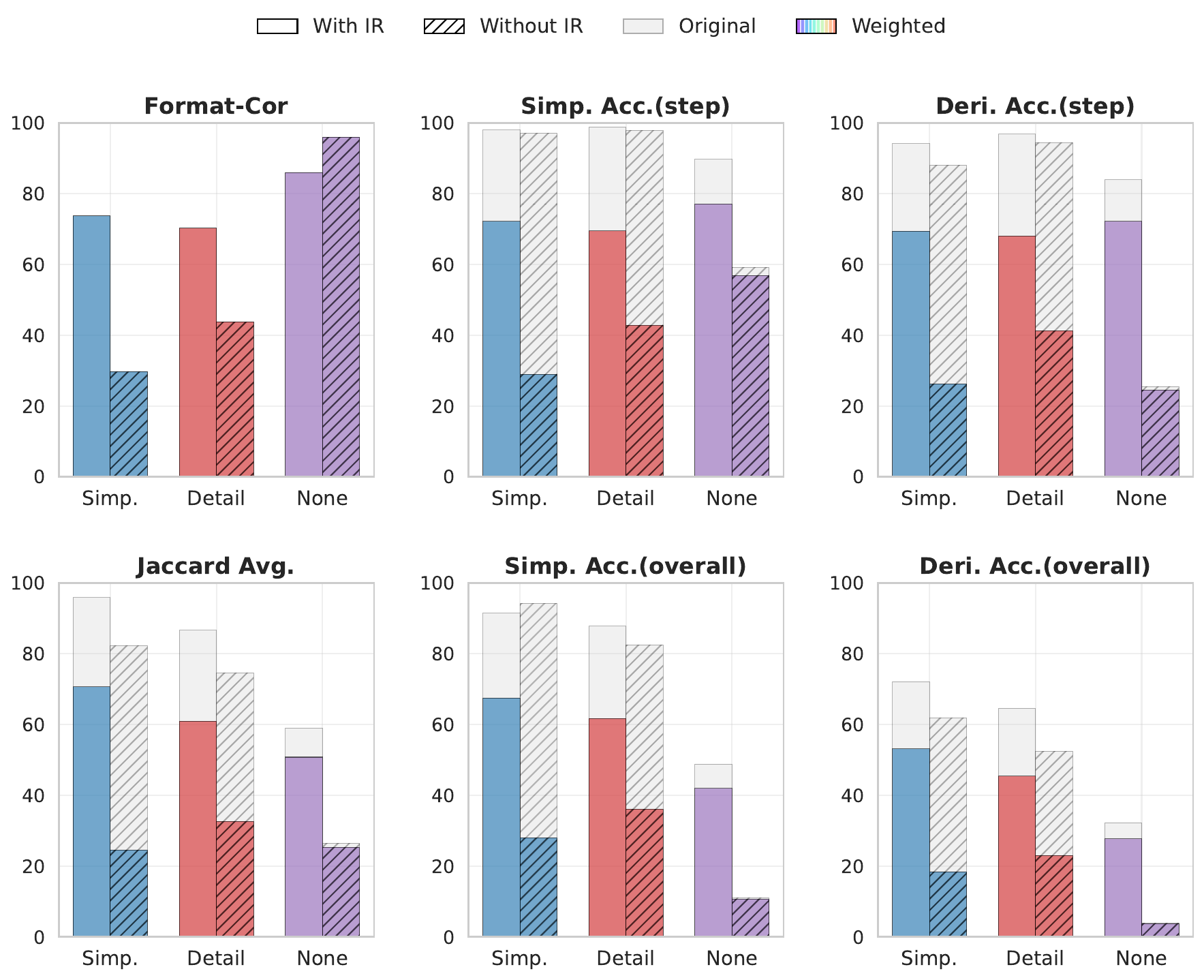}
        \caption{SNOMED CT}
        \label{fig:exp_Qwen3-32B-dataset-snomedct}
    \end{subfigure}
    \hfill
    \begin{subfigure}[t]{0.32\linewidth}
        \centering
        \includegraphics[width=\linewidth]{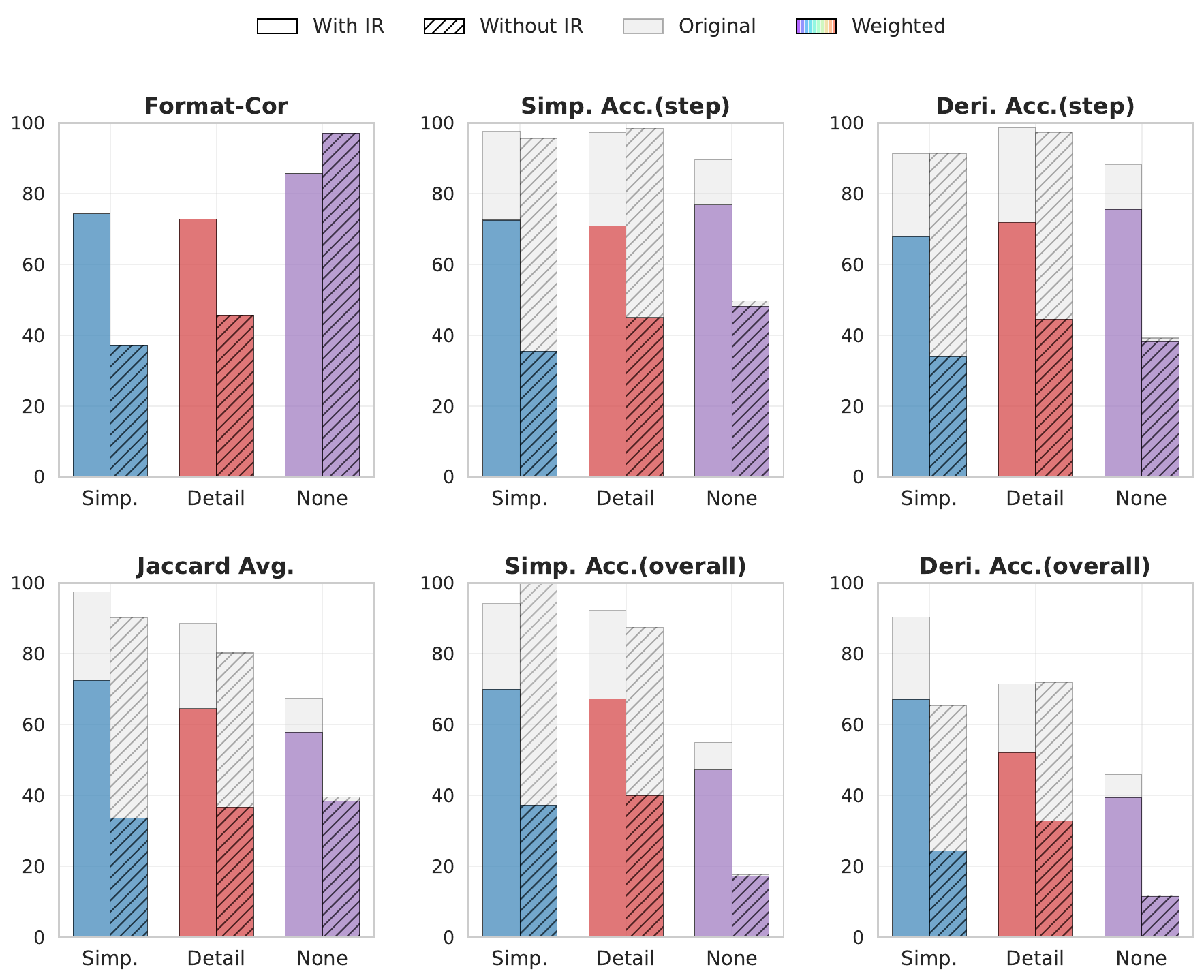}
        \caption{FoodOn}
    \end{subfigure}
    \hfill
    \begin{subfigure}[t]{0.32\linewidth}
        \centering
        \includegraphics[width=\linewidth]{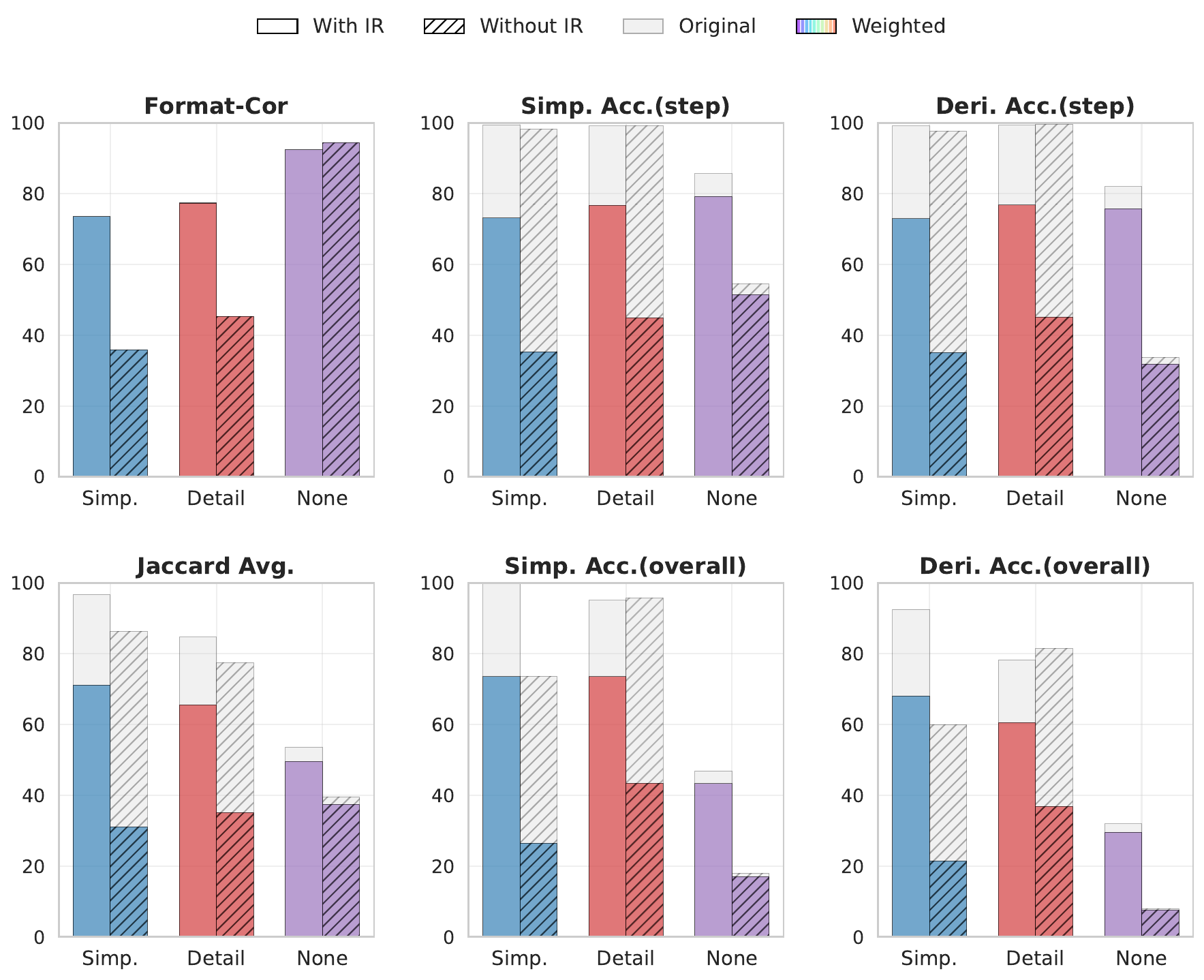}
        \caption{GO-Plus}
    \end{subfigure}

    \caption{Performance of Qwen3-32B on different datasets with different example in the prompt. \textit{Weighted} values are in color.}
    \label{fig:exp_Qwen3-32B-dataset}
\end{figure*}




\begin{figure*}[t]
    \centering
    \begin{subfigure}[t]{0.32\linewidth}
        \centering
        \includegraphics[width=\linewidth]{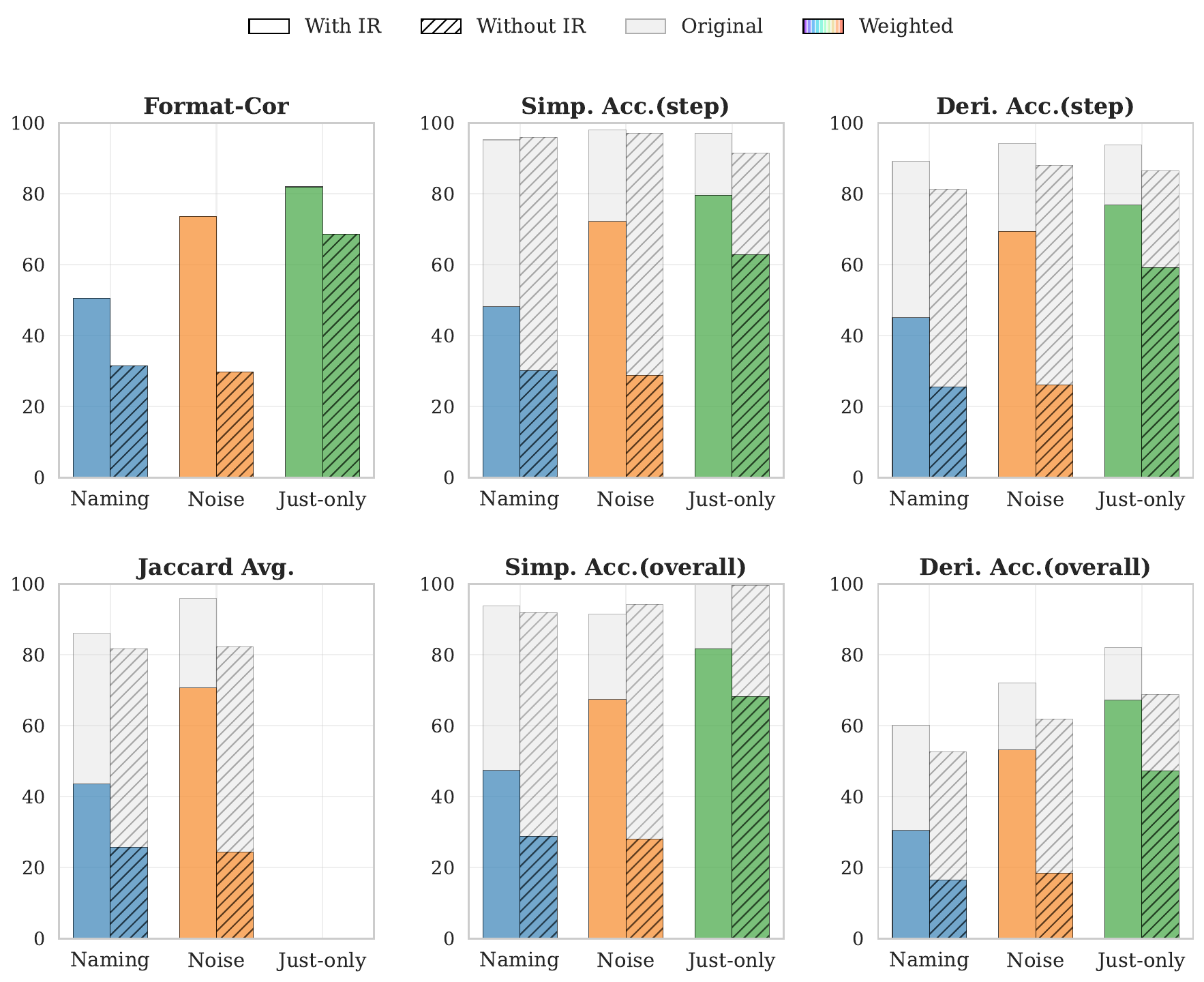}
        \caption{SNOMED CT}
           \label{fig:data_Qwen3-32B-dataset-snomedct}
    \end{subfigure}
    \hfill
    \begin{subfigure}[t]{0.32\linewidth}
        \centering
        \includegraphics[width=\linewidth]{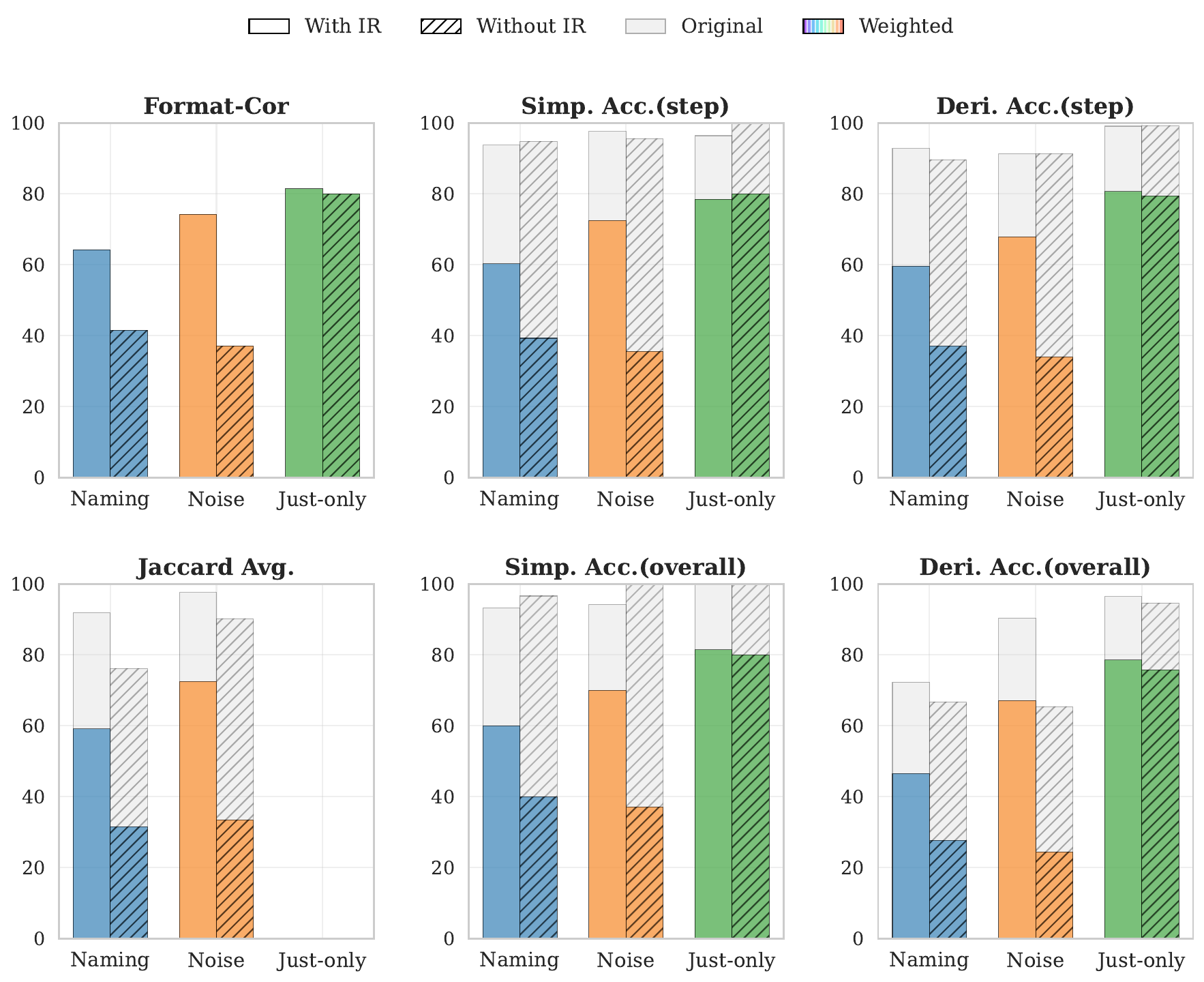}
        \caption{FoodOn}
    \end{subfigure}
    \hfill
    \begin{subfigure}[t]{0.32\linewidth}
        \centering
        \includegraphics[width=\linewidth]{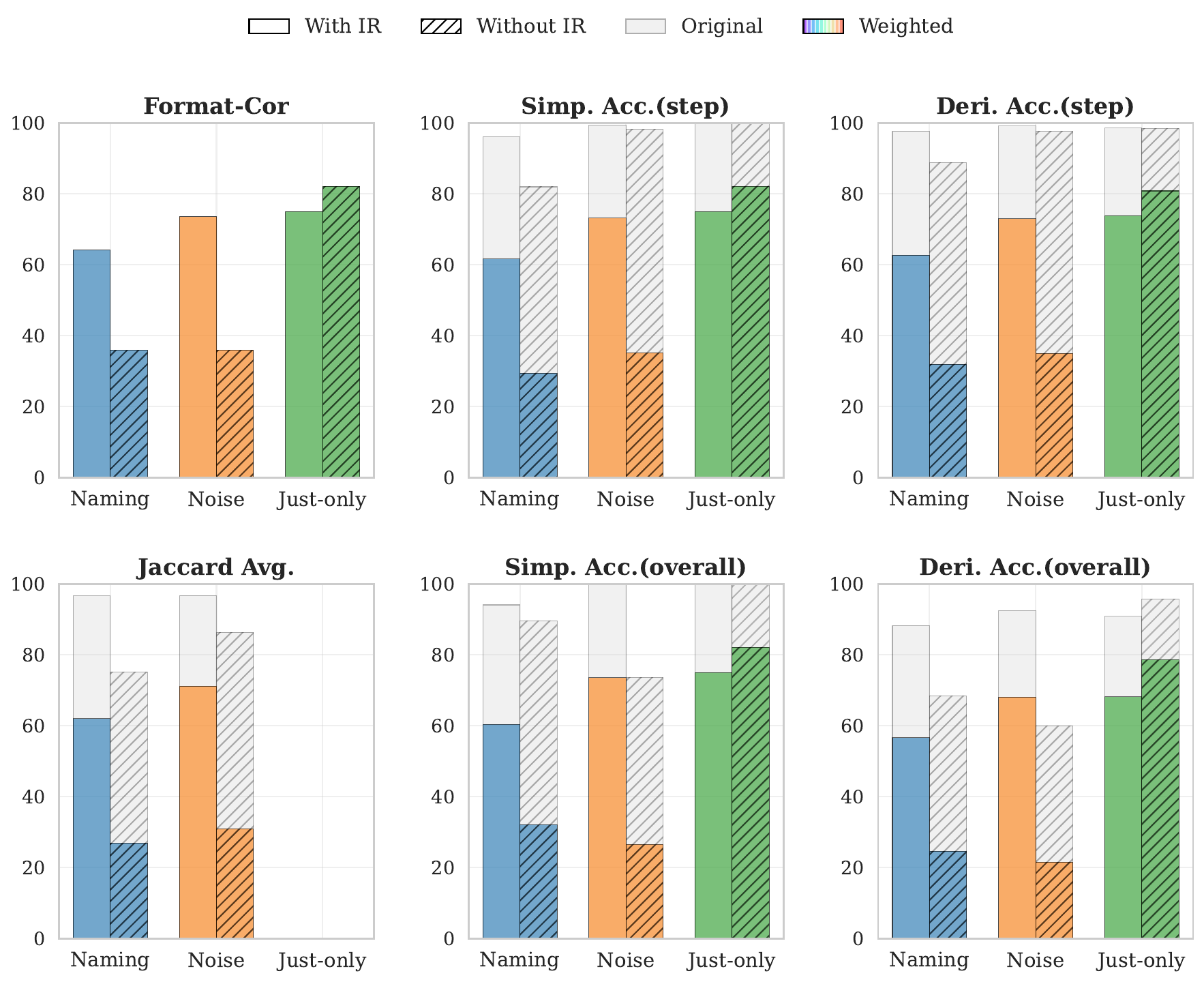}
        \caption{GO-Plus}
    \end{subfigure}

    \caption{Performance of Qwen3-32B on different datasets under different input data modes (with standard example in the prompt).}
    \label{fig:data_Qwen3-32B-dataset}
\end{figure*}

We examine two aspects to analyze its sensitivity of prompt to the example and input axioms as follows.  We focus on Qwen3-32B as the primary case study. 

\paragraph{Example modes.} We compare three modes: 
\begin{itemize}
    \item adding the standard example (\textbf{Simp.}),
    \item a version with detailed derivation  (\textbf{Detail}; see Figure~\ref{fig:prompt-example-detail}),
    \item no example (\textbf{None}),
\end{itemize}  
The main results for Qwen3-32B on SNOMED CT are presented in Figure~\ref{fig:exp_Qwen3-32B-dataset-snomedct}. Omitting examples increases format correctness but leads to substantial drops in result accuracy. In contrast, when no inference rules are provided, detailed examples consistently improve all performance metrics. However, when inference rules are included, providing detailed examples can slightly reduce result accuracy, likely due to the longer context, and the fact that the additional information in the detailed examples is already captured by the inference rules.

\paragraph{Input modes.} We evaluate three modes: 
\begin{itemize}
    \item the default \textbf{noise} setting with a balanced mix of true and noisy axioms,
    \item  the \textbf{just-only} setting where only true axioms are included, 
    \item  the \textbf{naming} setting where natural-language names or descriptions are added alongside axioms (e.g., $A:$ \textit{name/description of } $A$). 
\end{itemize}
Results on SNOMED CT (Figure~\ref{fig:data_Qwen3-32B-dataset-snomedct}) show that the \textit{just-only} mode consistently improves performance across all tasks, which is as expected since all noise axioms have been removed. In contrast, the \textit{naming} mode provides no improvements in the absence of inference rules and usually decreases performance when rules are included, indicating that although linguistic annotations could enhance performance, they are less effective than explicit logical rules in this case. 
Note that in \textit{just-only} mode, the Jaccard average does not make sense, since all axioms are justifications; therefore, we omit it from the figures.

\begin{figure*}
    \centering
    \begin{subfigure}[t]{0.48\linewidth}
        \centering
        \includegraphics[width=\linewidth]{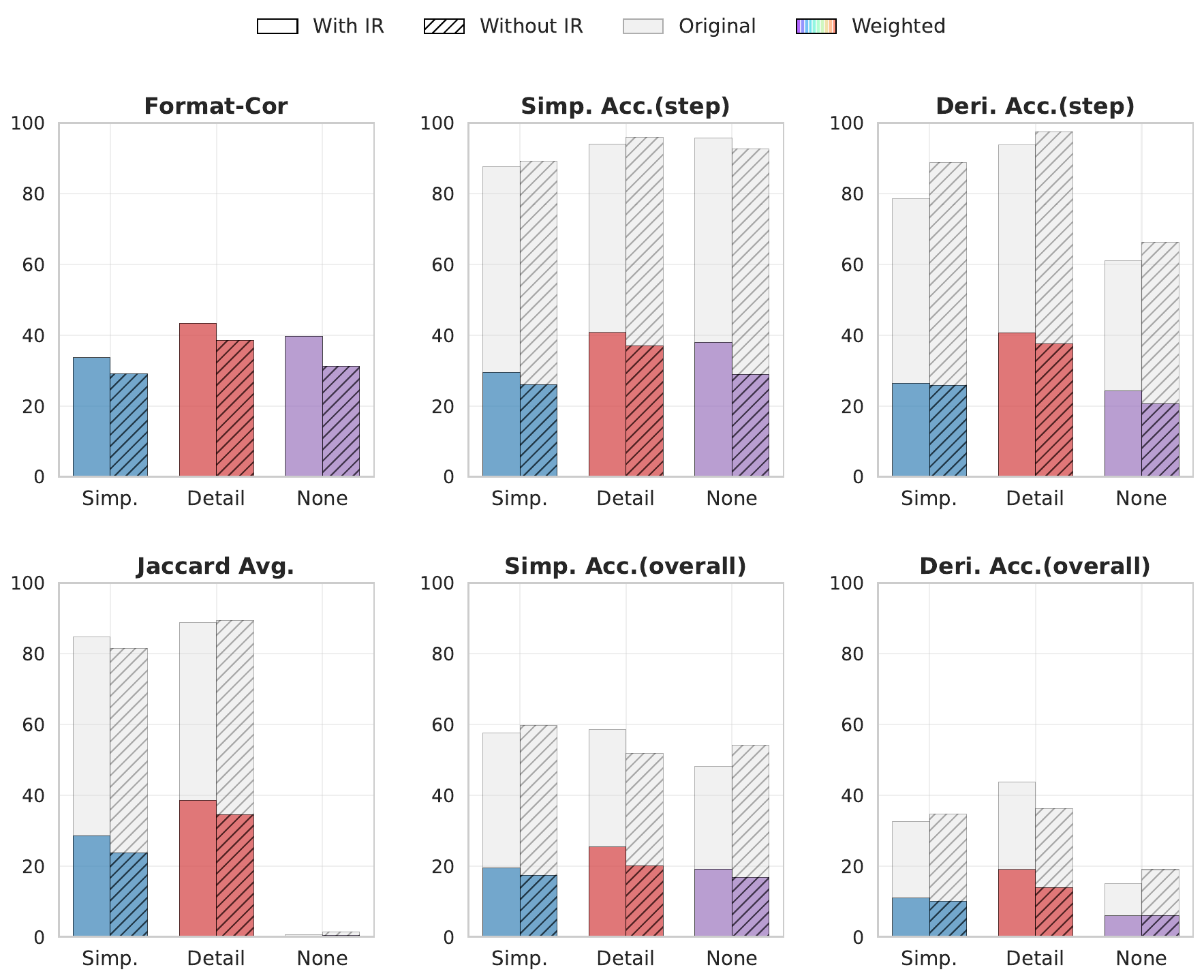}
        \caption{Example mode}
    \end{subfigure}
            \hfill
    \begin{subfigure}[t]{0.48\linewidth}
        \centering
        \includegraphics[width=\linewidth]{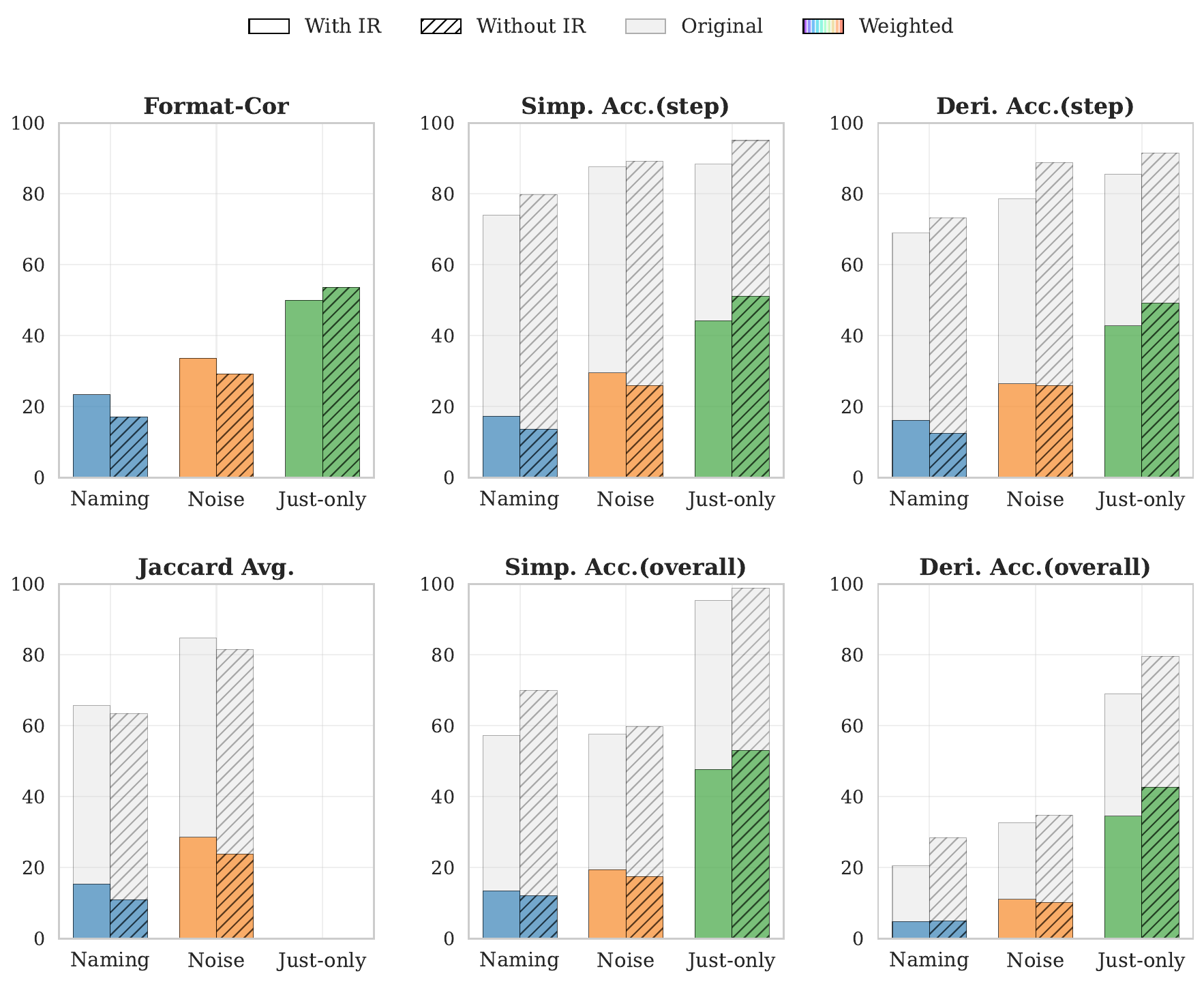}
        \caption{Data mode }
    \end{subfigure}
    \caption{Performance of Mistral-Small on SNOMED CT under different data and example modes.}
    \label{fig:data_mistral_small}
\end{figure*}

\begin{figure*}
    \centering
    \begin{subfigure}[t]{0.48\linewidth}
        \centering
        \includegraphics[width=\linewidth]{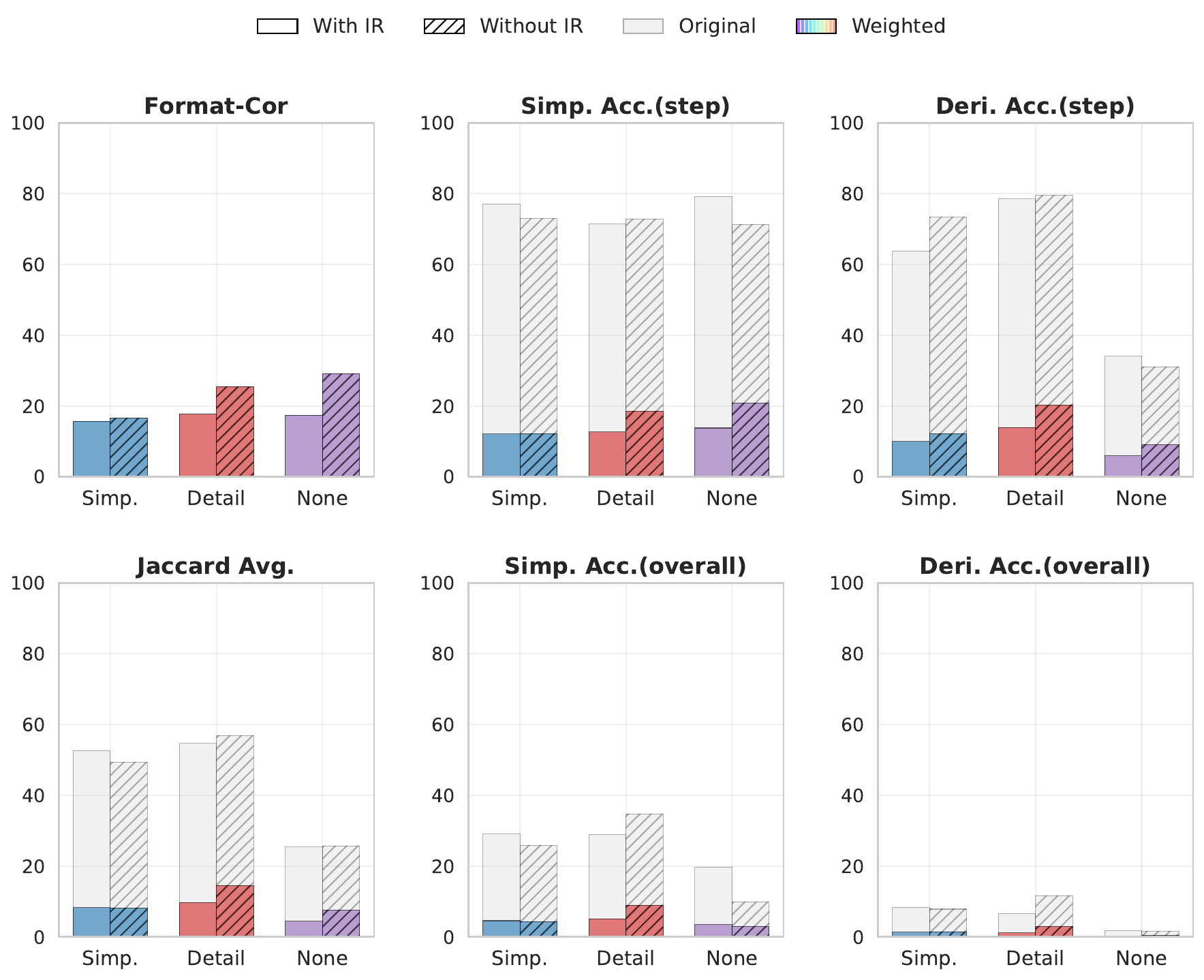}
        \caption{Example mode }
    \end{subfigure}
           \hfill
    \begin{subfigure}[t]{0.48\linewidth}
        \centering
        \includegraphics[width=\linewidth]{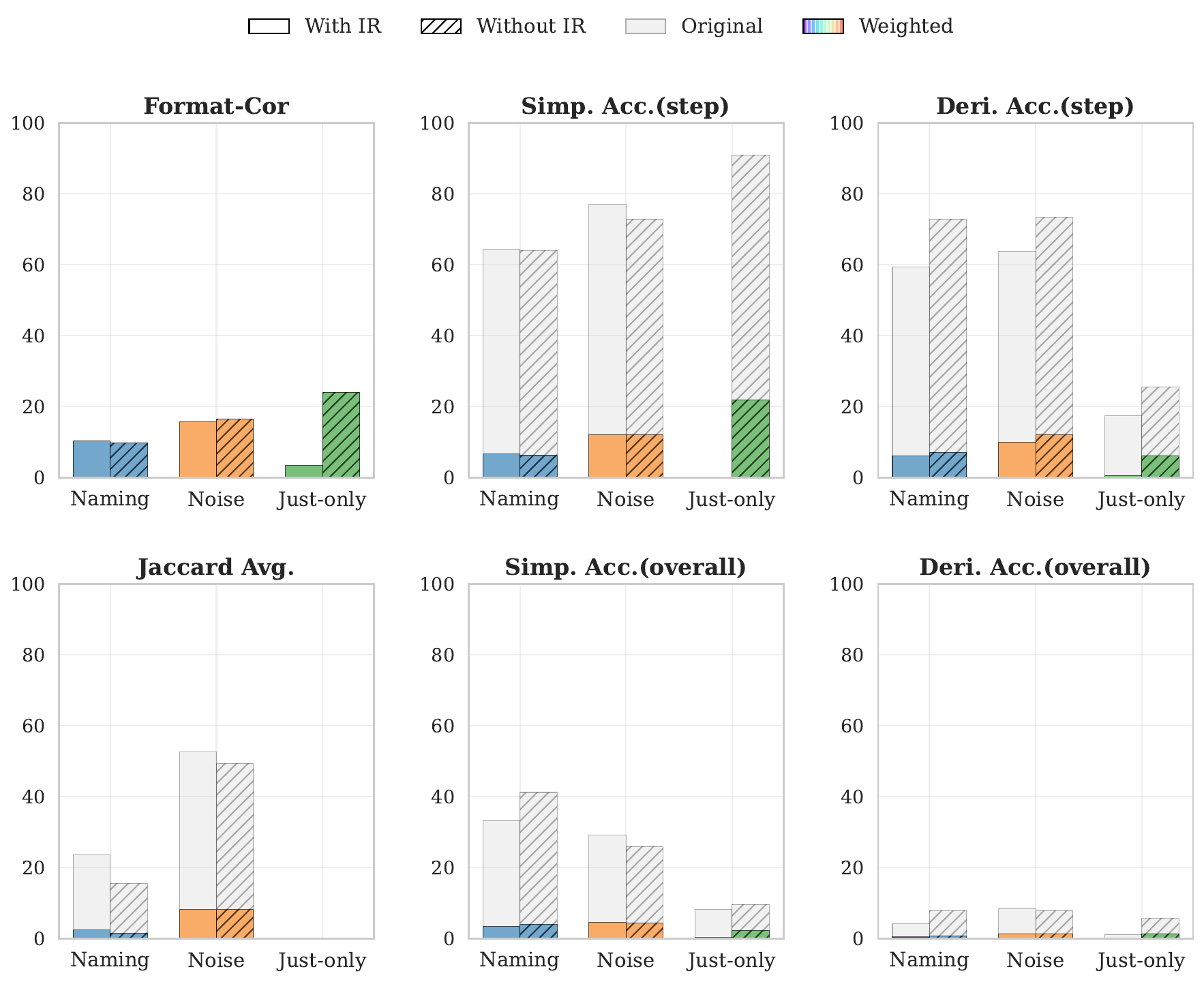}
        \caption{Data mode }
    \end{subfigure}
    \caption{Performance of DeepSeek-R1-Qwen-8B on SNOMED CT under different data and example modes.}
    \label{fig:data_r1-Qwen-8B}
\end{figure*}

The illustration of other models like DeepSeek-R1-Qwen-8B and Mistral-Small can be found in Figures \ref{fig:data_mistral_small} and \ref{fig:data_r1-Qwen-8B}, respectively.



\section{Case Studies}

In this section, we present several case studies covering both simple and complex scenarios. All samples are obtained using Qwen3-32B with examples and inference rules provided in the prompt, and extracted from Snomed CT. 

\begin{figure*}
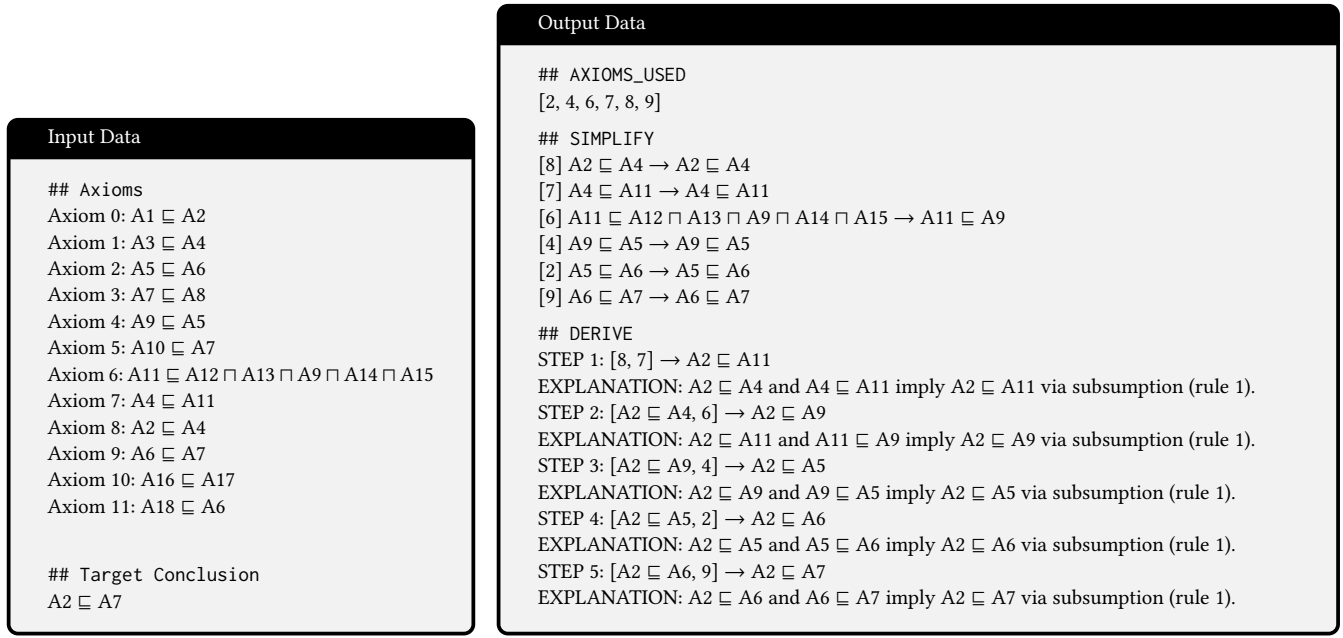

    \centering
    \small
    \begin{subfigure}{0.35\textwidth}
        \begin{tcolorbox}[title=Input Data,colback=lightgray,colframe=black]
        \texttt{\#\# Axioms}

        Axiom 0: A1 $\sqsubseteq$ A2

        Axiom 1: A3 $\sqsubseteq$ A4

        Axiom 2: A5 $\sqsubseteq$ A6

        Axiom 3: A7 $\sqsubseteq$ A8

        Axiom 4: A9 $\sqsubseteq$ A5

        Axiom 5: A10 $\sqsubseteq$ A7

        Axiom 6: A11 $\sqsubseteq$ A12 $\sqcap$ A13 $\sqcap$ A9 $\sqcap$ A14 $\sqcap$ A15

        Axiom 7: A4 $\sqsubseteq$ A11

        Axiom 8: A2 $\sqsubseteq$ A4

        Axiom 9: A6 $\sqsubseteq$ A7

        Axiom 10: A16 $\sqsubseteq$ A17

        Axiom 11: A18 $\sqsubseteq$ A6

        \vspace{2em}
        \texttt{\#\# Target Conclusion}

        A2 $\sqsubseteq$ A7
        \end{tcolorbox}
        \caption{Input axioms and target conclusion}
        \label{fig:input-box}
    \end{subfigure}
    \hfill
    \begin{subfigure}{0.63\textwidth}
        \vtop{\vskip0pt
        \begin{tcolorbox}[title=Output Data,colback=lightgray,colframe=black]
        \texttt{\#\# AXIOMS\_USED}

        [2, 4, 6, 7, 8, 9]

        \vspace{0.5em}
        \texttt{\#\# SIMPLIFY}

        [8] A2 $\sqsubseteq$ A4 $\rightarrow$ A2 $\sqsubseteq$ A4

        [7] A4 $\sqsubseteq$ A11 $\rightarrow$ A4 $\sqsubseteq$ A11

        [6] A11 $\sqsubseteq$ A12 $\sqcap$ A13 $\sqcap$ A9 $\sqcap$ A14 $\sqcap$ A15 $\rightarrow$ A11 $\sqsubseteq$ A9

        [4] A9 $\sqsubseteq$ A5 $\rightarrow$ A9 $\sqsubseteq$ A5

        [2] A5 $\sqsubseteq$ A6 $\rightarrow$ A5 $\sqsubseteq$ A6

        [9] A6 $\sqsubseteq$ A7 $\rightarrow$ A6 $\sqsubseteq$ A7

        \vspace{0.5em}
        \texttt{\#\# DERIVE}

        STEP 1: [8, 7] $\rightarrow$ A2 $\sqsubseteq$ A11 \\
        EXPLANATION: A2 $\sqsubseteq$ A4 and A4 $\sqsubseteq$ A11 imply A2 $\sqsubseteq$ A11 via subsumption (rule 1).

        STEP 2: [A2 $\sqsubseteq$ A4, 6] $\rightarrow$ A2 $\sqsubseteq$ A9 \\
        EXPLANATION: A2 $\sqsubseteq$ A11 and A11 $\sqsubseteq$ A9 imply A2 $\sqsubseteq$ A9 via subsumption (rule 1).

        STEP 3: [A2 $\sqsubseteq$ A9, 4] $\rightarrow$ A2 $\sqsubseteq$ A5 \\
        EXPLANATION: A2 $\sqsubseteq$ A9 and A9 $\sqsubseteq$ A5 imply A2 $\sqsubseteq$ A5 via subsumption (rule 1).

        STEP 4: [A2 $\sqsubseteq$ A5, 2] $\rightarrow$ A2 $\sqsubseteq$ A6 \\
        EXPLANATION: A2 $\sqsubseteq$ A5 and A5 $\sqsubseteq$ A6 imply A2 $\sqsubseteq$ A6 via subsumption (rule 1).

        STEP 5: [A2 $\sqsubseteq$ A6, 9] $\rightarrow$ A2 $\sqsubseteq$ A7 \\
        EXPLANATION: A2 $\sqsubseteq$ A6 and A6 $\sqsubseteq$ A7 imply A2 $\sqsubseteq$ A7 via subsumption (rule 1).
        \end{tcolorbox}}
        \caption{Output axioms used, simplification, and derivation}
        \label{fig:output-box}
    \end{subfigure}
    \vspace{-0.35cm}
\caption{A simple example with Atomic distance 6 and Justification size 6} 
\label{fig:case-simple}
\end{figure*}

In the relatively simple case shown in Figure~\ref{fig:case-simple}, most input axioms are atomic subsumptions, except for axiom 6. In this example, both the atomic distance to the given target and the justification size are 6. The LLM accurately selects the relevant justification axioms, as shown in the \texttt{AXIOMS\_USED} section, and correctly simplifies axiom 6 by preserving only the essential part, $A_{11}\sqsubseteq A_9$. Furthermore, the explanation provided in the \texttt{DERIVE} section is fully correct.

\begin{figure*}[ht]
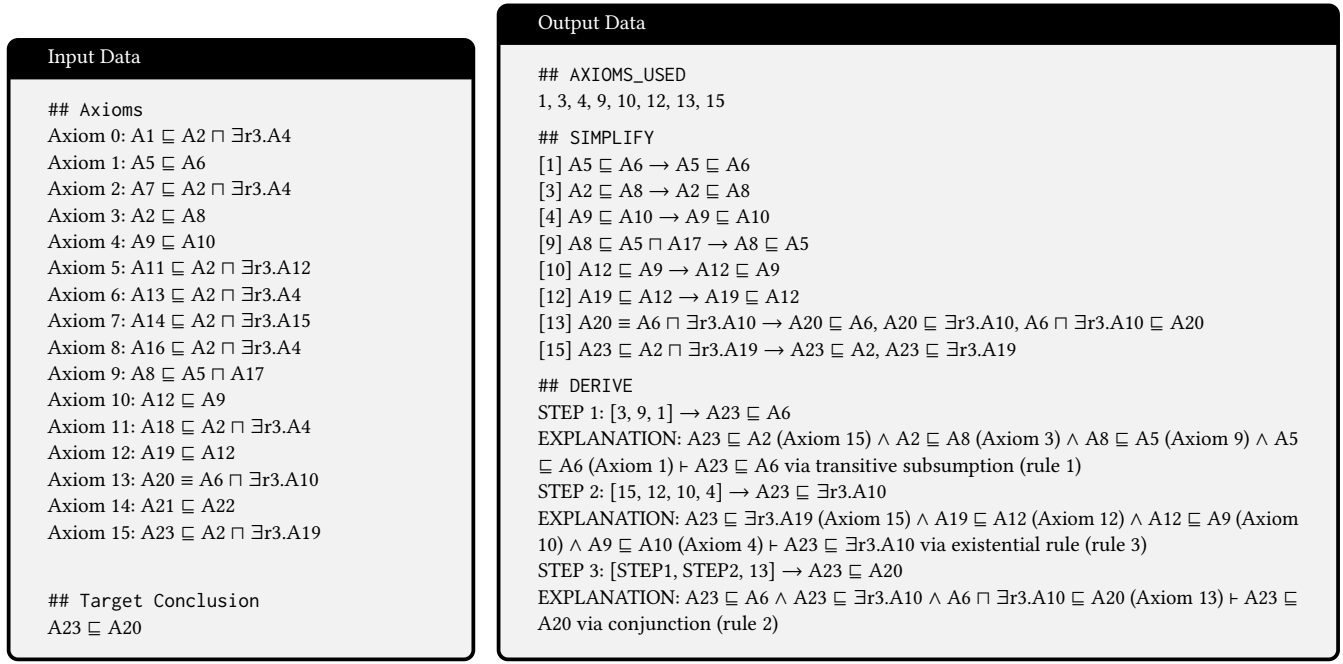

    \centering
    \small
    \begin{subfigure}{0.35\textwidth}
        \vtop{\vskip0pt
        \begin{tcolorbox}[title=Input Data,colback=lightgray,colframe=black]
        \texttt{\#\# Axioms}

        Axiom 0: A1 $\sqsubseteq$ A2 $\sqcap$ $\exists$r3.A4

        Axiom 1: A5 $\sqsubseteq$ A6

        Axiom 2: A7 $\sqsubseteq$ A2 $\sqcap$ $\exists$r3.A4

        Axiom 3: A2 $\sqsubseteq$ A8

        Axiom 4: A9 $\sqsubseteq$ A10

        Axiom 5: A11 $\sqsubseteq$ A2 $\sqcap$ $\exists$r3.A12

        Axiom 6: A13 $\sqsubseteq$ A2 $\sqcap$ $\exists$r3.A4

        Axiom 7: A14 $\sqsubseteq$ A2 $\sqcap$ $\exists$r3.A15

        Axiom 8: A16 $\sqsubseteq$ A2 $\sqcap$ $\exists$r3.A4

        Axiom 9: A8 $\sqsubseteq$ A5 $\sqcap$ A17

        Axiom 10: A12 $\sqsubseteq$ A9

        Axiom 11: A18 $\sqsubseteq$ A2 $\sqcap$ $\exists$r3.A4

        Axiom 12: A19 $\sqsubseteq$ A12

        Axiom 13: A20 $\equiv$ A6 $\sqcap$ $\exists$r3.A10

        Axiom 14: A21 $\sqsubseteq$ A22

        Axiom 15: A23 $\sqsubseteq$ A2 $\sqcap$ $\exists$r3.A19

        \vspace{2em}
        \texttt{\#\# Target Conclusion}

        A23 $\sqsubseteq$ A20
        \end{tcolorbox}}
        \caption{Input axioms and target conclusion}
        \label{fig:complex-input}
    \end{subfigure}
    \hfill
    \begin{subfigure}{0.63\textwidth}
        \vtop{\vskip0pt
        \begin{tcolorbox}[title=Output Data,colback=lightgray,colframe=black]
        \texttt{\#\# AXIOMS\_USED}

        1, 3, 4, 9, 10, 12, 13, 15

        \vspace{0.5em}
        \texttt{\#\# SIMPLIFY}

        [1] A5 $\sqsubseteq$ A6 $\rightarrow$ A5 $\sqsubseteq$ A6

        [3] A2 $\sqsubseteq$ A8 $\rightarrow$ A2 $\sqsubseteq$ A8

        [4] A9 $\sqsubseteq$ A10 $\rightarrow$ A9 $\sqsubseteq$ A10

        [9] A8 $\sqsubseteq$ A5 $\sqcap$ A17 $\rightarrow$ A8 $\sqsubseteq$ A5

        [10] A12 $\sqsubseteq$ A9 $\rightarrow$ A12 $\sqsubseteq$ A9

        [12] A19 $\sqsubseteq$ A12 $\rightarrow$ A19 $\sqsubseteq$ A12

        [13] A20 $\equiv$ A6 $\sqcap$ $\exists$r3.A10 $\rightarrow$  A20 $\sqsubseteq$ A6, A20 $\sqsubseteq$ $\exists$r3.A10, A6 $\sqcap$ $\exists$r3.A10 $\sqsubseteq$ A20

        [15] A23 $\sqsubseteq$ A2 $\sqcap$ $\exists$r3.A19 $\rightarrow$  A23 $\sqsubseteq$ A2, A23 $\sqsubseteq$ $\exists$r3.A19

        \vspace{0.5em}
        \texttt{\#\# DERIVE}

        STEP 1: [3, 9, 1] $\rightarrow$ A23 $\sqsubseteq$ A6 \\
        EXPLANATION: A23 $\sqsubseteq$ A2 (Axiom 15) $\wedge$ A2 $\sqsubseteq$ A8 (Axiom 3) $\wedge$ A8 $\sqsubseteq$ A5 (Axiom 9) $\wedge$ A5 $\sqsubseteq$ A6 (Axiom 1) $\vdash$ A23 $\sqsubseteq$ A6 via transitive subsumption (rule 1)

        STEP 2: [15, 12, 10, 4] $\rightarrow$ A23 $\sqsubseteq$ $\exists$r3.A10 \\
        EXPLANATION: A23 $\sqsubseteq$ $\exists$r3.A19 (Axiom 15) $\wedge$ A19 $\sqsubseteq$ A12 (Axiom 12) $\wedge$ A12 $\sqsubseteq$ A9 (Axiom 10) $\wedge$ A9 $\sqsubseteq$ A10 (Axiom 4) $\vdash$ A23 $\sqsubseteq$ $\exists$r3.A10 via existential rule (rule 3)

        STEP 3: [STEP1, STEP2, 13] $\rightarrow$ A23 $\sqsubseteq$ A20 \\
        EXPLANATION: A23 $\sqsubseteq$ A6 $\wedge$ A23 $\sqsubseteq$ $\exists$r3.A10 $\wedge$ A6 $\sqcap$ $\exists$r3.A10 $\sqsubseteq$ A20 (Axiom 13) $\vdash$ A23 $\sqsubseteq$ A20 via conjunction (rule 2)
        \end{tcolorbox}}
        \caption{Output axioms used, simplification, and derivation}
        \label{fig:complex-output}
    \end{subfigure}
    \vspace{-0.35cm}
    \caption{A complex example with Atomic distance 4 and Justification size 8, split into input and output boxes}
    \label{fig:case-complex-split}
\end{figure*}

Figure~\ref{fig:case-complex-split} illustrates a more complex case, where the input axioms include compound expressions such as $A_{20} \equiv A_6 \sqcap \exists r_3.A_{10}$. Deriving the target conclusion requires inference patterns beyond simple transitive subsumptions. In this scenario, the LLM correctly identifies the justification axioms, but the simplification process is less satisfactory: axioms 13 and 15 are split into simpler forms, yet not all non-essential parts are removed. For instance, only the component A6 $A_6 \sqcap \exists r_3.A_{10} \sqsubseteq A_{20}$ from axiom 13 is required. Additionally, the derivation process contains errors. In the first derivation step, one cannot derive $A_{23} \sqsubseteq A_6$ from axioms 3, 9, and 1. The correct derivation steps are:

\begin{itemize}
    \item[\textbf{Step1}] [Axioms 15, 3, 9, 1] $\vdash A_{23} \sqsubseteq A_6$
    \item[\textbf{Step2}]  [Axioms 12, 10, 4] $\vdash A_{19} \sqsubseteq A_{10}$
    \item[\textbf{Step3}]  [Step 2, Axiom 15] $\vdash A_{23} \sqsubseteq \exists r_3. A_{10}$
    \item[\textbf{Step4}]  [Step 1, Step 3, Axiom 13] $\vdash A_{23} \sqsubseteq A_{20}$
\end{itemize}

Although the \texttt{DERIVE} output is almost correct, it has the small mistake: the omission of axiom 15 in the premise set for Step 1, despite being mentioned in the explanation.

Finally, Figure~\ref{fig:case-hard} presents an even more complex example. In this case, the axioms are highly nested and intricate, and the LLM is unable to produce the result within the token limits.

\begin{figure*}
    \centering
    \small
    \begin{tcolorbox}[title=Reasoning for Target Conclusion,colback=lightgray,colframe=black]
    \texttt{\#\# Axioms}
    
    \texttt{Axiom 0: } A1 $\equiv$ A2 $\sqcap$ $\exists$r3.( $\exists$r4.A5 $\sqcap$ $\exists$r6.A7 ) $\sqcap$ $\exists$r3.( $\exists$r4.A5 $\sqcap$ $\exists$r6.A8 )
    
    \texttt{Axiom 1: } A9 $\sqsubseteq$ A10 $\sqcap$ A11
    
    \texttt{Axiom 2: } A12 $\equiv$ A13 $\sqcap$ $\exists$r3.$\exists$r6.A14
    
    \texttt{Axiom 3: } A15 $\equiv$ A2 $\sqcap$ $\exists$r3.( $\exists$r4.A5 $\sqcap$ $\exists$r6.A16 )
    
    \texttt{Axiom 4: } A17 $\equiv$ A2 $\sqcap$ $\exists$r3.( $\exists$r4.A5 $\sqcap$ $\exists$r6.A18 )
    
    \texttt{Axiom 5: } A19 $\sqsubseteq$ A20 $\sqcap$ $\exists$r3.$\exists$r6.A21
    
    \texttt{Axiom 6: } A22 $\equiv$ A23 $\sqcap$ $\exists$r24.A25
    
    \texttt{Axiom 7: } A26 $\sqsubseteq$ A9 $\sqcap$ A27
    
    \texttt{Axiom 8: } A28 $\equiv$ A2 $\sqcap$ $\exists$r3.( $\exists$r4.A5 $\sqcap$ $\exists$r6.A18 ) $\sqcap$ $\exists$r3.( $\exists$r4.A5 $\sqcap$ $\exists$r6.A29 )
    
    \texttt{Axiom 9: } A30 $\sqsubseteq$ A2 $\sqcap$ $\exists$r3.$\exists$r6.A21
    
    \texttt{Axiom 10: } A31 $\equiv$ A2 $\sqcap$ $\exists$r3.( $\exists$r4.A5 $\sqcap$ $\exists$r6.A32 )
    
    \texttt{Axiom 11: } A33 $\sqsubseteq$ A2
    
    \texttt{Axiom 12: } A34 $\sqsubseteq$ A35 $\sqcap$ A36
    
    \texttt{Axiom 13: } A37 $\equiv$ A2 $\sqcap$ $\exists$r3.( $\exists$r4.A5 $\sqcap$ $\exists$r6.A38 ) $\sqcap$ $\exists$r3.( $\exists$r4.A5 $\sqcap$ $\exists$r6.A22 )
    
    \texttt{Axiom 14: } A39 $\equiv$ A2 $\sqcap$ $\exists$r3.( $\exists$r4.A5 $\sqcap$ $\exists$r6.A29 )
    
    \texttt{Axiom 15: } A20 $\sqsubseteq$ A13
    
    \texttt{Axiom 16: } A40 $\equiv$ A20 $\sqcap$ $\exists$r3.( $\exists$r4.A41 $\sqcap$ $\exists$r6.A42 ) $\sqcap$ $\exists$r3.( $\exists$r4.A41 $\sqcap$ $\exists$r6.A43 )
    
    \texttt{Axiom 17: } A23 $\sqsubseteq$ A34 $\sqcap$ A16
    
    \texttt{Axiom 18: } A44 $\sqsubseteq$ A45 $\sqcap$ A26 $\sqcap$ A46
    
    \texttt{Axiom 19: } A36 $\sqsubseteq$ A44
    
    \texttt{Axiom 20: } A2 $\sqsubseteq$ A47 $\sqcap$ A19 $\sqcap$ $\exists$r3.$\exists$r6.A21
    
    \texttt{Axiom 21: } A48 $\equiv$ A2 $\sqcap$ $\exists$r3.( $\exists$r4.A5 $\sqcap$ $\exists$r6.A8 )
    
    \texttt{Axiom 22: } A49 $\sqsubseteq$ A14
    
    \texttt{Axiom 23: } A50 $\sqsubseteq$ A51 $\sqcap$ A52
    
    \texttt{Axiom 24: } A53 $\sqsubseteq$ A49 $\sqcap$ A54
    
    \texttt{Axiom 25: } A55 $\sqsubseteq$ A56 $\sqcap$ $\exists$r3.( $\exists$r4.A57 $\sqcap$ $\exists$r6.A21 )
    
    \texttt{Axiom 26: } A58 $\equiv$ A20 $\sqcap$ $\exists$r3.( $\exists$r4.A59 $\sqcap$ $\exists$r60.A61 $\sqcap$ $\exists$r6.A62 $\sqcap$ $\exists$r63.A64 )
    
    \texttt{Axiom 27: } A65 $\equiv$ A2 $\sqcap$ $\exists$r3.( $\exists$r4.A66 $\sqcap$ $\exists$r6.A44 )
    
    \texttt{Axiom 28: } A67 $\equiv$ A2 $\sqcap$ $\exists$r3.( $\exists$r4.A5 $\sqcap$ $\exists$r6.A23 )
    
    \texttt{Axiom 29: } A10 $\sqsubseteq$ A53
    
    \vspace{0.5em}
    
    \texttt{\#\# Target Conclusion}
    
    \texttt{A37 }$\sqsubseteq$\texttt{ A12}
    
\end{tcolorbox}
    \vspace{-0.35cm}
    \caption{A hard example with atomic distance 8 and justification size 15.}
    \label{fig:case-hard}
\end{figure*}

\end{document}

%% file: marcos.tex
\newcommand{\R}{\ensuremath{\mathbb{R}}\xspace}
\newcommand{\Sp}{\ensuremath{\mathbb{S}}\xspace}
\newcommand{\M}{\ensuremath{\mathcal{M}}\xspace}
\newcommand{\C}{\ensuremath{\mathcal{C}}\xspace}
\newcommand{\N}{\ensuremath{\widetilde{\mathcal{N}}}\xspace}
\newcommand{\Nstar}{\ensuremath{\mathcal{N}^*}\xspace}
\newcommand{\Q}{\ensuremath{\mathcal{Q}}\xspace}

\newcommand{\x}{\ensuremath{\mathbf{x}}\xspace}
\newcommand{\bc}{\ensuremath{\mathbf{c}}\xspace}
\newcommand{\bo}{\ensuremath{\mathbf{o}}\xspace}
\newcommand{\bR}{\ensuremath{\mathbf{R}}\xspace}
\newcommand{\bB}{\ensuremath{\mathbf{B}}\xspace}
\newcommand{\bK}{\ensuremath{\mathbf{K}}\xspace}
\newcommand{\ba}{\ensuremath{\mathbf{a}}\xspace}
\newcommand{\bP}{\ensuremath{\mathbf{P}}\xspace}
\newcommand{\Op}{\ensuremath{\mathbf{O}}\xspace}
\newcommand{\y}{\ensuremath{\mathbf{y}}\xspace}
\newcommand{\V}{\ensuremath{\mathbf{V}}\xspace}
\newcommand{\z}{\ensuremath{\mathbf{z}}\xspace}
\newcommand{\A}{\ensuremath{\mathbf{A}}\xspace}
\newcommand{\X}{\ensuremath{\mathbf{X}}\xspace}
\newcommand{\I}{\ensuremath{\mathbf{I}}\xspace}
\newcommand{\D}{\ensuremath{\mathbf{D}}\xspace}
\newcommand{\bL}{\ensuremath{\mathbf{L}}\xspace}
\newcommand{\Lmc}{\ensuremath{\mathcal{L}}\xspace}
\newcommand{\Cmc}{\ensuremath{\mathcal{C}}\xspace}
\newcommand{\W}{\ensuremath{\mathbf{W}}\xspace}
\newcommand{\bH}{\ensuremath{\mathbf{H}}\xspace}
\newcommand{\Wm}{\ensuremath{\mathbf{U}}\xspace}
\newcommand{\Ori}{\ensuremath{\mathbf{o}}\xspace}
\newcommand{\tran}{\ensuremath{\mathit{\tran}}}
\newcommand{\Bbox}{\ensuremath{\mathit{Box}}}
\newcommand{\Vol}{\ensuremath{\mathit{Vol}}}
\newcommand{\Agg}{\textit{\Agg}}
\newcommand{\subdist}{atomic distance\xspace}

\newcommand{\sig}{\ensuremath{\textit{sig}}\xspace}

\newcommand{\tA}{\ensuremath{\Tilde{\mathbf{A}}}\xspace}
\newcommand{\tD}{\ensuremath{\Tilde{\mathbf{D}}}\xspace}
\newcommand{\bb}{\ensuremath{\mathbf{b}}\xspace}
\newcommand{\bu}{\ensuremath{\mathbf{u}}\xspace}
\newcommand{\bv}{\ensuremath{\mathbf{v}}\xspace}
\newcommand{\bw}{\ensuremath{\mathbf{u}}\xspace}
\newcommand{\m}{\ensuremath{\mathbf{m}}\xspace}
\newcommand{\bh}{\ensuremath{\mathbf{h}}\xspace}
\newcommand{\br}{\ensuremath{\mathbf{r}}\xspace}
\newcommand{\bt}{\ensuremath{\mathbf{t}}\xspace}
\newcommand{\bk}{\ensuremath{\mathbf{k}}\xspace}
\newcommand{\be}{\ensuremath{\mathbf{e}}\xspace}
\newcommand{\bp}{\ensuremath{\mathbf{p}}\xspace}

\newcommand{\boxsqel}{Box$^2$EL\xspace}

\newcommand{\Bmc}{\ensuremath{\mathcal{B}}\xspace}
\newcommand{\Omc}{\ensuremath{\mathcal{O}}\xspace}
\newcommand{\ORI}{\ensuremath{\mathcal{O}_{RI}}\xspace}
\newcommand{\OCI}{\ensuremath{\mathcal{O}_{CI}}\xspace}
\newcommand{\Imc}{\ensuremath{\mathcal{I}}\xspace}
\newcommand{\Jmc}{\ensuremath{\mathcal{J}}\xspace}
\newcommand{\Xmc}{\ensuremath{\mathcal{X}}\xspace}
\newcommand{\Mmc}{\ensuremath{\mathcal{M}}\xspace}
\newcommand{\starM}{$\top\!\bot^\ast$-module}
\newcommand{\ALC}{\ensuremath{\mathcal{ALC}}\xspace}
\newcommand{\ALCH}{\ensuremath{\mathcal{ALCH}}\xspace}
\newcommand{\EL}{\ensuremath{\mathcal{EL}}\xspace}
\newcommand{\ELp}{\ensuremath{\mathcal{EL}^+}\xspace}
\newcommand{\ELH}{\ensuremath{\mathcal{ELH}}\xspace}

\newcommand{\NC}{\ensuremath{\mathsf{N_C}}\xspace}
\newcommand{\NR}{\ensuremath{\mathsf{N_R}}\xspace}
\newcommand{\NI}{\ensuremath{\mathsf{N_I}}\xspace}
\newcommand{\ND}{\ensuremath{\mathsf{N_D}}\xspace}

\newcommand{\bmu}{\boldsymbol{\mu}}